\pdfoutput=1

\documentclass[11pt]{article}

\usepackage[]{EMNLP2022}

\usepackage{times}
\usepackage{latexsym}
\usepackage{CJKutf8}
\usepackage{amsmath}
\usepackage{verbatim}
\usepackage{booktabs}
\usepackage{amssymb}
\usepackage{pifont}
\usepackage{stfloats}
\usepackage{multicol}
\usepackage{float}
\usepackage{graphicx}
\usepackage{subfigure}
\usepackage{multirow}
\usepackage{array}
\newcommand{\PreserveBackslash}[1]{\let\temp=\\#1\let\\=\temp}

\newcolumntype{C}[1]{>{\PreserveBackslash\centering}p{#1}}
\newcolumntype{R}[1]{>{\PreserveBackslash\raggedleft}p{#1}}
\newcolumntype{L}[1]{>{\PreserveBackslash\raggedright}p{#1}}
\usepackage{diagbox}
\usepackage[noend]{algpseudocode}
\usepackage{booktabs}
\usepackage{amsfonts}
\usepackage{bbm}
\usepackage{mathtools}

\usepackage{xcolor}
\usepackage[linesnumbered,ruled,vlined]{algorithm2e}

\SetCommentSty{mycommfont}

\SetKwInput{KwInput}{Input}                
\SetKwInput{KwOutput}{Output}              
\SetKwInput{KwInit}{Init} 

\makeatletter
\def\hlinew#1{%
  \noalign{\ifnum0=`}\fi\hrule \@height #1 \futurelet
   \reserved@a\@xhline}
\makeatother

\usepackage[T1]{fontenc}

\usepackage[utf8]{inputenc}

\usepackage{microtype}

\title{Wait-info Policy: Balancing Source and Target at Information Level \\ for Simultaneous Machine Translation}

\author{Shaolei Zhang \textsuperscript{\rm 1,2},
Shoutao Guo \textsuperscript{\rm 1,2},
    Yang Feng \textsuperscript{\rm 1,2}\thanks{ \ \ Corresponding author: Yang Feng.} \\
        \textsuperscript{\rm 1}{Key Laboratory of Intelligent Information Processing} \\ Institute of Computing Technology, Chinese Academy of Sciences (ICT/CAS) \\
    { \textsuperscript{\rm 2} {University of Chinese Academy of Sciences, Beijing, China}} \\
     \texttt{\{\href{mailto:zhangshaolei20z@ict.ac.cn}{zhangshaolei20z}, 
     \href{mailto:guoshoutao22z@ict.ac.cn}{guoshoutao22z},
     \href{mailto:fengyang@ict.ac.cn}{fengyang}\}@ict.ac.cn}  }

\begin{document}
\maketitle
\begin{abstract}
Simultaneous machine translation (SiMT) outputs the translation while receiving the source inputs, and hence needs to balance the received source information and translated target information to make a reasonable decision between waiting for inputs or outputting translation. Previous methods always balance source and target information at the token level, either directly waiting for a fixed number of tokens or adjusting the waiting based on the current token. In this paper, we propose a \emph{Wait-info Policy} to balance source and target at the information level. We first quantify the amount of information contained in each token, named \emph{info}. Then during simultaneous translation, the decision of waiting or outputting is made based on the comparison results between the total info of previous target outputs and received source inputs. Experiments show that our method outperforms strong baselines under and achieves better balance via the proposed info\footnote{Code is available at \url{https://github.com/ictnlp/Wait-info}}.
\end{abstract}

\section{Introduction}
Simultaneous machine translation (SiMT) \cite{Cho2016,gu-etal-2017-learning,ma-etal-2019-stacl} outputs the translation while receiving the source sentence, aiming at the trade-off between translation quality and latency.
Therefore, a policy is required for SiMT to decide between waiting for the source inputs (i.e., READ) or outputting translations (i.e., WRITE), the core of which is to wisely balance the received source information and the translated target information. When the source information is less, the model should wait for more inputs for a high-quality translation; conversely, when the translated target information is less, the model should output translations for a low latency.

\begin{figure}[t]
\centering
\subfigure[Wait-k policy: treats each token equally, and lags $k$ tokens.]{
\includegraphics[width=3in]{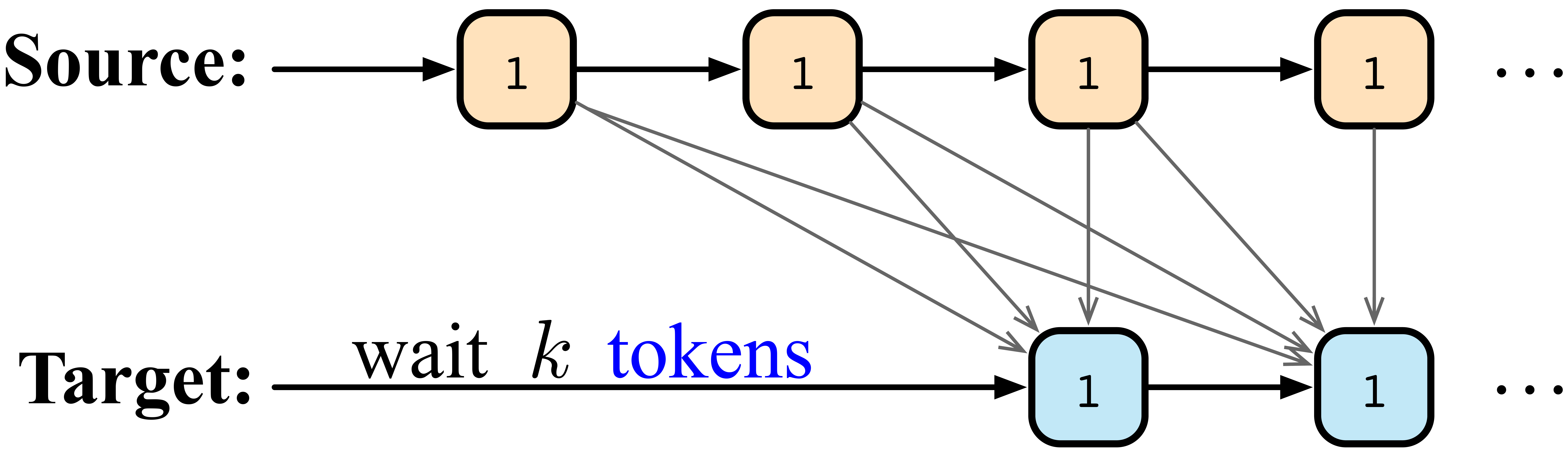} \label{ill1}
}
\subfigure[Wait-info policy: quantifies the information in each token, named \emph{info} (e.g., $0.5,1.7,\cdots$), and keeps the target information always less than the received source information $\mathcal{K}$ info.]{
\includegraphics[width=3in]{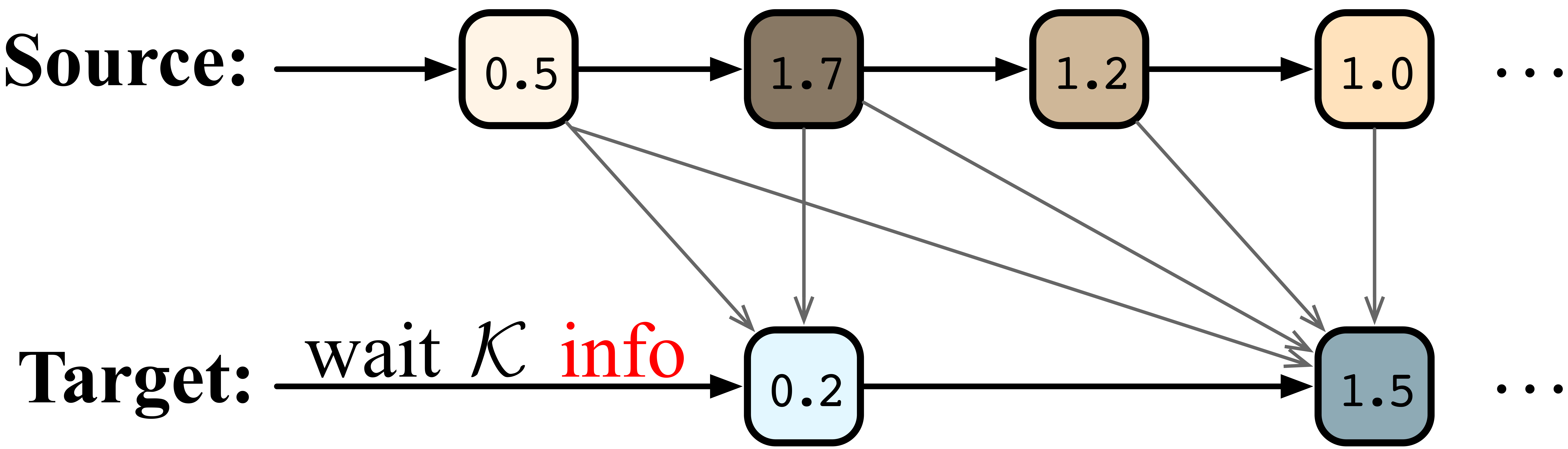} \label{ill2}
}
\caption{Schematic diagram of Wait-info v.s. Wait-k.}
\label{ill}
\end{figure}
Existing SiMT policies, involving fixed and adaptive, always balance source and target at the token level, i.e., treating each source and target token equally when determining READ/WRITE. Fixed policies decide READ/WRITE based on the number of received source tokens \cite{ma-etal-2019-stacl,zhang-feng-2021-universal}, such as wait-k policy \cite{ma-etal-2019-stacl} simply considers each source token to be equivalent and lets the target outputs always lag the source inputs by $k$ tokens, as shown in Figure \ref{ill1}. Fixed policies are always limited by the fact that the policy cannot be adjusted according to complex inputs, making them difficult to get the best trade-off. Adaptive policies predict READ/WRITE according to the current source and target tokens \cite{Arivazhagan2019,Ma2019a} and thereby get a better trade-off, but they often ignore and under-utilize the difference between tokens when deciding READ/WRITE. Besides, existing adaptive policies always rely on complicated training \cite{Ma2019a,miao-etal-2021-generative} or additional labeled data \cite{Zheng2019b,zhang-etal-2020-learning-adaptive,alinejad-etal-2021-translation}, making them more computationally expensive than fixed policies. 

Treating each token equally when balancing source and target is not the optimal choice for SiMT policy. Many studies have shown that different words have significantly different functions in translation \cite{lin-etal-2018-learning,moradi-etal-2019-interrogating,chen-etal-2020-content}, often divided into content words (i.e., noun, verb, $\cdots$) and function words (i.e., conjunction, preposition, $\cdots$), where the former express more important meaning and the latter is less informative. Accordingly, tokens with different amounts of information should also play different roles in the SiMT policy, where more informative tokens should play a more dominant role because they bring more information to SiMT model \cite{gma,ITST}. Therefore, explicitly differentiating various tokens rather than treating them equally when determining READ/WRITE will be beneficial to developing a more precise SiMT policy.

In this paper, we differentiate various source and target tokens based on the amount of information they contain, aiming to balance received source information and translated target information at the information level. To this end, we propose \emph{wait-info policy}, a simple yet effective policy for SiMT. As shown in Figure \ref{ill2}, we first quantify the amount of information contained in each token through a scalar, named \emph{info}, which is jointly learned with the attention mechanism in an unsupervised manner. During the simultaneous translation, READ/WRITE decisions are made by balancing the total info of translated target information and received source information. If the received source information is more than translated target information by $\mathcal{K}$ info or more, the model outputs translation, otherwise the model waits for the next input. Experiments and analyses show that our method outperforms strong baselines and effectively quantifies the information contained in each token.

\section{Related Work}
\textbf{SiMT Policy}\quad Recent policies fall into fixed and adaptive. For fixed policy, \citet {ma-etal-2019-stacl} proposed wait-k policy, which first READ $k$ source tokens and then READ/WRITE one token alternately. \citet{multipath} proposed an efficient multi-path training for wait-k policy to randomly sample $k$ during training. \citet{future-guided} proposed future-guide training for wait-k policy, which introduces a full-sentence MT to guide training. \citet{zhang-feng-2021-icts} proposed a char-level wait-k policy. \citet{zhang-feng-2021-universal} proposed a mixture-of-experts wait-k policy to develop a universal SiMT model. For adaptive policy, \citet {gu-etal-2017-learning} trained an agent to decide READ/WRITE via reinforcement learning. \citet {Arivazhagan2019} proposed MILk, which predicts a Bernoulli variable to determine READ/WRITE. \citet {Ma2019a} proposed MMA to implement MILk on Transformer. \cite{dualpath} proposed dual-path SiMT to enhance MMA with dual learning.
\citet{zheng-etal-2020-simultaneous} developed adaptive wait-k through heuristic ensemble of multiple wait-k models. \citet{miao-etal-2021-generative} proposed a generative framework to generate READ/WRITE decisions. \citet{gma} proposed Gaussian multi-head attention to decide READ/WRITE based on alignments.

Previous policies always treat each token equally when determining READ/WRITE, ignoring the fact that tokens with different amounts of information often play different roles in SiMT policy. Our method aims to develop a more precise SiMT policy by differentiating the importance of various tokens when determining READ/WRITE.

\textbf{Information Modeling in NMT}\quad Linguistics divides words into content words and function words according to their information and functions in the sentence. Therefore, modeling the information contained in each word is often used to improve the NMT performance. \citet{moradi-etal-2019-interrogating} and \citet{chen-etal-2020-content} used the word frequency to indicate how much information each word contains, and the words with lower frequencies contain more information. \citet{liu-etal-2020-norm} and \citet{kobayashi-etal-2020-attention} found that the norm of word embedding is related to the token information in NMT. \citet{lin-etal-2018-learning} and \citet{zhang-feng-2021-modeling-concentrated} argued that the attention mechanism for different types of word should be different, where the attention distribution of content word tends to be more concentrated. 

Our method explores the usefulness of modeling information for SiMT policy, and proposes an unsupervised method to quantify the information of tokens through the attention mechanism, achieving good explainability.

\section{Background}
\textbf{Full-sentence MT}\quad For a translation task, we denote the source sentence as $\textbf{x}\!=\!\left ( x_{1} , \cdots ,x_{n}\right )$ with source length $n$ and the target sentence as $\textbf{y}\!=\!\left ( y_{1} , \cdots ,y_{m}\right )$ with target length $m$. Transformer \cite{NIPS2017_7181} is the most widely used architecture for full-sentence MT, consisting of an encoder and a decoder. Encoder maps $\textbf{x}$ to source hidden states $\textbf{z}\!=\!\left ( z_{1} , \cdots ,z_{n}\right )$. Decoder maps $\textbf{y}$ to target hidden states $\textbf{s}\!=\!\left ( s_{1} , \cdots ,s_{m}\right )$, and then performs translating. Specifically, each encoder layer contains two sub-layers: self-attention and feed-forward network (FFN), while each decoder layer contains three sub-layers: self-attention, cross-attention and FFN. 
Both self-attention and cross-attention are implemented through the dot-product attention between query $\mathbf{Q}$ and key $\mathbf{K}$, calculated as:
\begin{align}
    e _{ij}= &\;\frac{Q_{i}W^{Q}\left ( K_{j}W^{K} \right )^{\top}}{\sqrt{d_{k}}},   \label{eq0} \\
    \alpha _{ij}= &\;\mathrm{softmax}\left (e _{ij}  \right). \label{eq1}
\end{align}
where $e _{ij}$ is the similarity score between $Q_{i}$ and $K_{j}$, and $\alpha _{ij}$ is the normalized attention weight. $d_{k}$ is the input dimension, $W^{Q}$ and $W^{K}$ are projection parameters.  
More specifically, self-attention extracts the monolingual representation of source or target tokens, so the query and key both come from the source hidden states $\textbf{z}$ or target hidden states $\textbf{s}$. While cross-attention extracts the cross-lingual representation through measuring the correlation between target and source token, so query comes from the target hidden states $\textbf{s}$, and key comes from the source hidden states $\textbf{z}$.

\textbf{Wait-k Policy}\quad Simultaneous machine translation (SiMT) determines when to start translating each target token through a policy. Wait-k policy \cite{ma-etal-2019-stacl} is the most widely used policy for SiMT, which refers to first waiting for $k$ source tokens and then translating and waiting for one token alternately, i.e., the target outputs always lagging $k$ tokens behind the source inputs. Formally, when translating $y_{i}$, wait-k policy forces the SiMT model to wait for $g_{k}\left(i\right)$ source tokens, where $g_{k}\left(i\right)$ is calculated as:
\begin{gather}
g_{k}\left(i\right)= \min\left\{k+i-1, n\right\}. \label{eq3}
\end{gather}

\section{Method}
\label{sec:method}

To differentiate various tokens when determining READ/WRITE, we quantify the amount of information contained in each source and target token, named info. As shown in Figure \ref{model}, we propose \emph{info-aware Transformer} to jointly learn the quantified info with the attention mechanism in an unsupervised manner. Then based on the quantified info, we propose \emph{wait-info policy} to balance the received source information and translated target information. The details are as follows.

\subsection{Info Quantification}

\begin{figure}[t]
\centering
\includegraphics[width=3in]{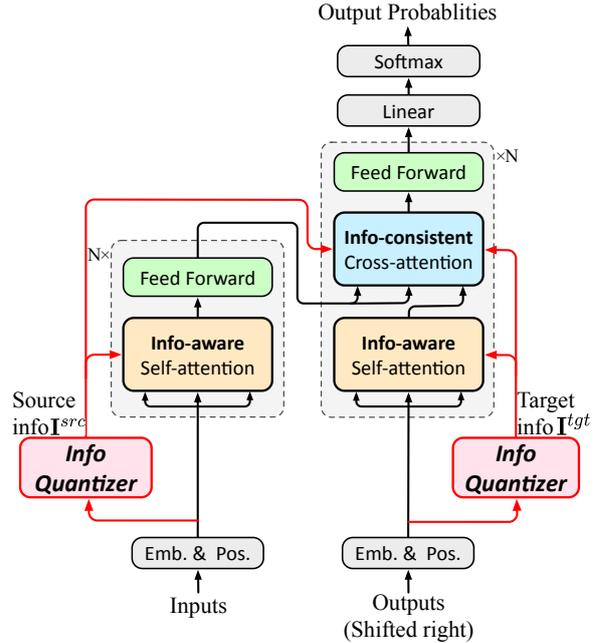}
\caption{Architecture of the proposed info-aware Transformer, where we omit residual connection and layer normalization in the figure for clarity.}
\label{model}
\end{figure}

To quantify the amount of information in each token, we use a scalar to represent how much information each token contains, named \emph{info}. We denote the info of the source tokens and the target tokens as $\mathbf{I}^{src}\! \in \!\mathbb{R}^{n\times 1}$ and $\mathbf{I}^{tgt} \!\in\! \mathbb{R}^{m\times 1}$, respectively, where $I^{src}_{j}$ and $I^{tgt}_{i}$ represent the info of $x_{j}$ and $y_{i}$, and the higher info means that the token has more information.

To predict $\mathbf{I}^{src}$ and $\mathbf{I}^{tgt}$, we introduce two \emph{Info Quantizer}s before the encoder and decoder to respectively quantify the information of each source and target token, as shown in Figure \ref{model}. Specifically, the info quantizer is implemented by a 3-layer feed-forward network (FFN):
\begin{align}
    \mathbf{I}^{src}=&\;2\times \mathrm{sigmoid}\left(\mathrm{FFN}\left(\mathbf{x}\right)\right), \label{eq4}\\
    \mathbf{I}^{tgt}=&\;2\times \mathrm{sigmoid}\left(\mathrm{FFN}\left(\mathbf{y}\right)\right).
\end{align}
For the formulation of the following wait-info policy, $2\!\times\!\mathrm{sigmoid}\!\left(\cdot\right)$ is used to restrict the quantified info $I^{src}_{j},I^{tgt}_{i} \!\in\! \left(0,2\right)$.

Further, in a translation task, source sentence and target sentence should be semantically equivalent \cite{finch-etal-2005-using,post-eval}, so the total information of source tokens should be equal to that of target tokens. To this end, we introduce an info-sum loss $\mathcal{L}_{sum}$ to constrain the total info of the source tokens and target tokens, calculated as:
\begin{gather}
    \mathcal{L}_{sum}=\left\|\sum_{j=1}^{n}I^{src}_{j}-\zeta   \right\|_{2}+ \left\|\sum_{i=1}^{m}I^{tgt}_{i}-\zeta   \right\|_{2}, \label{eq6}
\end{gather}
where $\zeta$ is a hyperparameter to represent the total info, and we set $\zeta\!=\!\frac{m+n}{2}$ (i.e., average length of source and target) to control the average info to be around 1. Therefore, the final loss $ \mathcal{L}$ is:
\begin{gather}
    \mathcal{L}=\mathcal{L}_{ce}+\lambda\mathcal{L}_{sum}, \label{eq7}
\end{gather}
where $\mathcal{L}_{ce}$ is the original cross-entropy loss for the translation \cite{NIPS2017_7181}. $\lambda$ is a hyperparameter and we set $\lambda=0.3$ in our experiments.

\subsection{Learning of Quantified Info}

The form of quantified info $\mathbf{I}^{src}$ and $\mathbf{I}^{tgt}$ has been constrained through Eq.(\ref{eq4}-\ref{eq7}), and then the key challenge is how to encourage the quantified info to accurately reflect the amount of information each token contains. Since the tokens with different amounts of information often show different preferences in the attention distribution \cite{lin-etal-2018-learning}, we propose an unsupervised method to learn the quantified info through the attention mechanism. As shown in Figure \ref{model}, we introduce an info-aware Transformer, consisting of \emph{info-aware self-attention} and \emph{info-consistent cross-attention}.

\textbf{Info-aware Self-attention}\quad
Self-attentions in both encoder and decoder are used to extract monolingual representations of tokens, where tokens with different amounts of information tend to exhibit different attention distributions \cite{lin-etal-2018-learning, zhang-feng-2021-modeling-concentrated}. Specifically, tokens with much information, such as content words, tend to pay more attention to themselves. For the tokens with less information, since they have less meaning in themselves, they need more context information and thereby pay less attention to themselves. Therefore, we use the quantified info to bias the tokens' attention to themselves, thereby encouraging those tokens that tend to focus more on themselves to get higher info. Specifically, based on the original self-attention in Eq.(\ref{eq0},\ref{eq1}), we add the quantified info $I^{\tau }_{i},\tau \!\in\!\left\{src,tgt\right\}$ (respectively used for encoder and decoder self-attention) on the token's similarity to itself $e_{ii}$ \cite{lin-etal-2018-learning}, and then normalize them with $\mathrm{softmax}\left(\cdot\right)$ to get the info-aware self-attention $\beta _{ij}$, calculated as:
\begin{align}
    \widetilde{e} _{ij}= &\; \begin{cases}
 e _{ij}+ \left(I^{\tau }_{i}-1\right),& \textrm{ if }\; i=j \\
e _{ij}\;\;\;\;\;\;\;\;\;\;\;\;\;\;\;\;\;\:, & \textrm{ otherwise }
\end{cases}, \\
\beta _{ij}= &\; \mathrm{softmax}\left (\widetilde{e} _{ij}  \right).
\end{align}
If $I^{\tau }_{i}\!>\!1$ (i.e., containing more information), the token will pay more attention to itself, otherwise the token will focus more on other tokens to extract context information. Therefore, the info can be learned from the attention distribution.

\textbf{Info-consistent Cross-attention}\quad 
In addition to modeling the token info in a monolingual context, the consistency of the token info between target and source is also crucial for the SiMT policy, which ensures that the received source information and the target information can be accurately balanced under the same criterion. For consistency, the target and source tokens with high similarity (i.e., those with high cross-attention scores) should have similar info. Therefore, we scale the cross-attention with the info consistency between target and source, where the info consistency is measured by $L_{1}$ distance between target and source info. Info-consistent cross-attention $\gamma _{ij}$ is calculated as:
\begin{align}
    \widetilde{\gamma} _{ij}= &\; \alpha _{ij}\times \left(2-\left|I^{tgt }_{i}-I^{src }_{j} \right|\right), \\
\gamma _{ij}= &\; \widetilde{\gamma} _{ij}/\sum _{j}\widetilde{\gamma} _{ij},
\end{align}
where $\left(2-\left|I^{tgt }_{i}-I^{src }_{j} \right|\right)\in \left(0,2\right]$ measures the info consistent between $y_{i}$ and $x_{j}$.

Overall, we apply the proposed info-aware self-attention $\beta _{ij}$ and info-consistent cross-attention $\gamma _{ij}$ to replace the original attention for the learning of the quantified info.

\begin{algorithm}[t]
\DontPrintSemicolon
  \KwInput{source inputs $\mathbf{x}$ (incremental), lagging info $\mathcal{K}$,\\ $\;\;\;\;\;\;\;\;\;\;\;\hat{y}_{0}=$ \textit{BeginOfSequence} }
  \KwOutput{target outputs $\mathbf{\hat{y}}$}
  \KwInit{target idx $i=1$, source idx $j=1$}
  \While{$\;\hat{y}_{i-1}\neq$ EndOfSequence}{
  
    Calculate info $I^{src}_{j}$ and $I^{tgt}_{i}$ \\
    \tcc{1) Source info is more;$\:$ or $\;\;\;\;\;\;\;\;\;\;\;\;\;\;\;$2) Inputs is complete.}
    \eIf(\tcp*[f]{WRITE}){$\;\;\sum_{l=1}^{j}\! I^{src}_{l}\; \geq\; \sum_{l=1}^{i}\! I^{tgt}_{l}\!+\!\mathcal{K}$$\;\;\;\mathbf{or}$\\$\;\;\;\;\;\;x_{j}==$ EndOfSequence \\$\!\!$  }{
      Translate $\hat{y}_{i}$ with $\left(x_{1},\cdots,x_{j}\right)$; \\
      $i\leftarrow i+1$;
    }
    (\tcp*[f]{READ})
    {
      Wait for next source input $x_{j+1}$; \\
      $j\leftarrow j+1$; \\
    }
    
    }
  \Return $\mathbf{\hat{y}}$;
\caption{Wait-info Policy}
\label{algor}
\end{algorithm}
\subsection{Wait-info Policy}

Owing to the quantification and learning of info, we get $\mathbf{I}^{src}$ and $\mathbf{I}^{tgt}$ to reflect how much information that source and target tokens contain. Then, we develop \emph{wait-info policy} for SiMT to balance source and target at the information level. 

Borrowing the idea from the wait-k policy that requires the target outputs to lag behind the source inputs by $k$ tokens \cite{ma-etal-2019-stacl}, wait-info policy keeps that the target information is always less than the received source information $\mathcal{K}$ info, where $\mathcal{K} $ is the lagging info, a hyperparameter to control the latency. Formally, we denote the number of source tokens that the SiMT model waits for before translating $y_{i}$ as $g_{\mathcal{K}}\!\left ( i \right )$, calculated as:
\begin{gather}
    g_{\mathcal{K}}\!\left ( i \right )=\underset{j}{\!\mathrm{argmin}}\!\left (  \sum_{l=1}^{j}\! I^{src}_{l} \geq \sum_{l=1}^{i}\! I^{tgt}_{l}\!+\!\mathcal{K}\right ). \label{eq11}
\end{gather}
The specific decoding process of wait-info policy is shown in Algorithm \ref{algor}.

During training, we mask out the source token $x_{j}$ that $j\!>\! g_{\mathcal{K}}\!\left ( i \right )$ to simulate the incomplete source sentence. Besides, we apply multi-path training \cite{multipath} to randomly sample different $\mathcal{K}$ in each batch to enhance the training efficiency.

\section{Experiment}

\subsection{Datasets}

\textbf{IWSLT15\footnote{\url{nlp.stanford.edu/projects/nmt/}} English $\!\rightarrow \!$ Vietnamese (En$\rightarrow$Vi)} (133K pairs) We use TED tst2012 (1553 pairs) as the dev set and TED tst2013 (1268 pairs) as the test set. Following the previous setting \cite{Ma2019a}, we replace tokens that frequency less than 5 by $\left \langle unk \right \rangle$, and the vocabulary sizes of English and Vietnamese are 17K and 7.7K respectively.

\textbf{WMT15\footnote{\url{www.statmt.org/wmt15/translation-task}} German $\!\rightarrow\! $ English (De$\rightarrow$En)} (4.5M pairs) We use newstest2013 (3000 pairs) as the dev set and newstest2015 (2169 pairs) as the test set. BPE \cite{sennrich-etal-2016-neural} is applied with 32K merge operations and the vocabulary is shared.

\subsection{System Settings}
We conduct experiments on following systems.

{\bf Full-sentence MT} Standard Transformer model \cite{NIPS2017_7181}, which waits for the complete source sentence and then starts translating.

{\bf Wait-k} Wait-k policy \cite{ma-etal-2019-stacl}, which first READ $k$ source tokens, and then alternately READ one token and WRITE one token.

{\bf Efficient Wait-k} An efficient multi-path training for wait-k \cite{multipath}, which randomly samples $k$ between batches during training.

{\bf Adaptive Wait-k} An adaptive policy via a heuristic composition of a set of wait-k models (e.g., $k$ from 1 to 13) \cite{zheng-etal-2020-simultaneous}. Adaptive Wait-k uses the tokens number of target and source to select a wait-k model to generate a target token, and then decides whether to output or not according to the generating probability.

{\bf MoE Wait-k\footnote{\url{github.com/ictnlp/MoE-Waitk}}} Mixture-of-experts wait-k policy \cite{zhang-feng-2021-universal}, which applies multiple experts to perform wait-k policy with various $k$ to consider the translation under multiple latency.

{\bf{MMA}\footnote{\url{github.com/pytorch/fairseq/tree/master/examples/simultaneous_translation}}} Monotonic multi-head attention (MMA) \cite{Ma2019a}, which uses a Bernoulli variable 0/1 to decide READ/WRITE and Bernoulli variable is jointly learning with multi-head attention.

{\bf GSiMT} Generative SiMT \cite{miao-etal-2021-generative}, which applies a generative framework to predict a Bernoulli variable to decide READ/WRITE, and uses the dynamic programming to train the policy.

{\bf GMA\footnote{\url{github.com/ictnlp/GMA}}} Gaussian multi-head attention (GMA) \cite{gma}, which uses a Gaussian prior to learn the alignments in attention, and then performs READ/WRITE based on the alignments.

{\bf {Wait-info}} The proposed method in Sec.\ref{sec:method}.

The implementation of all systems are based on Transformer \cite{NIPS2017_7181} and adapted from Fairseq Library \cite{ott-etal-2019-fairseq}. Following \citet{Ma2019a}, we apply Transformer-Small (4 heads) for En$\rightarrow$Vi, Transformer-Base (8 heads) and Transformer-Big (16 heads) for De$\rightarrow$En. Since GSiMT involves dynamic programming with expensive training costs, we only report GSiMT on De$\rightarrow$En with Transformer-Base, the same as its original setting \cite{miao-etal-2021-generative}.
For evaluation, we report BLEU \cite{papineni-etal-2002-bleu} for translation quality and Average Lagging (AL) \cite{ma-etal-2019-stacl} for latency.
Average lagging evaluates the number of tokens lagging behind the ideal policy, calculated as:
\begin{gather}
    \mathrm{AL}=\frac{1}{\tau }\sum_{i=1}^{\tau}g\left(i\right)-\frac{i-1}{m/n},
\end{gather}
where $\tau \!=\!\mathrm{argmax}_{i}\left ( g\left(i\right)\!=\! n\right )$, and $g\left(i\right)$ is number of waited source tokens before translating $y_{i}$.

\subsection{Main Results}

\begin{figure*}[t]
\centering
\subfigure[En$\rightarrow$Vi, Transformer-Small]{
\includegraphics[width=2in]{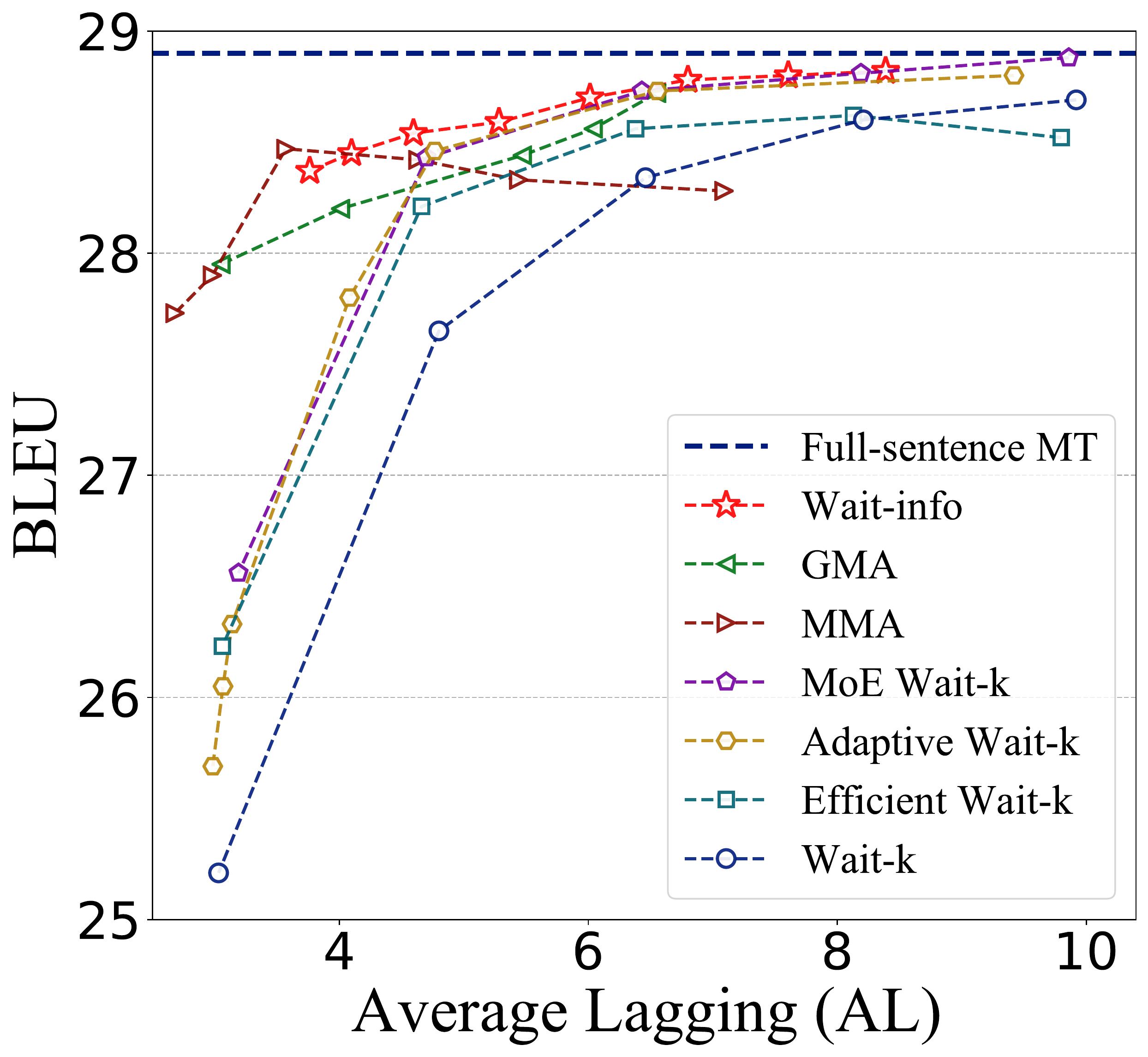}
}
\subfigure[De$\rightarrow$En, Transformer-Base]{
\includegraphics[width=2in]{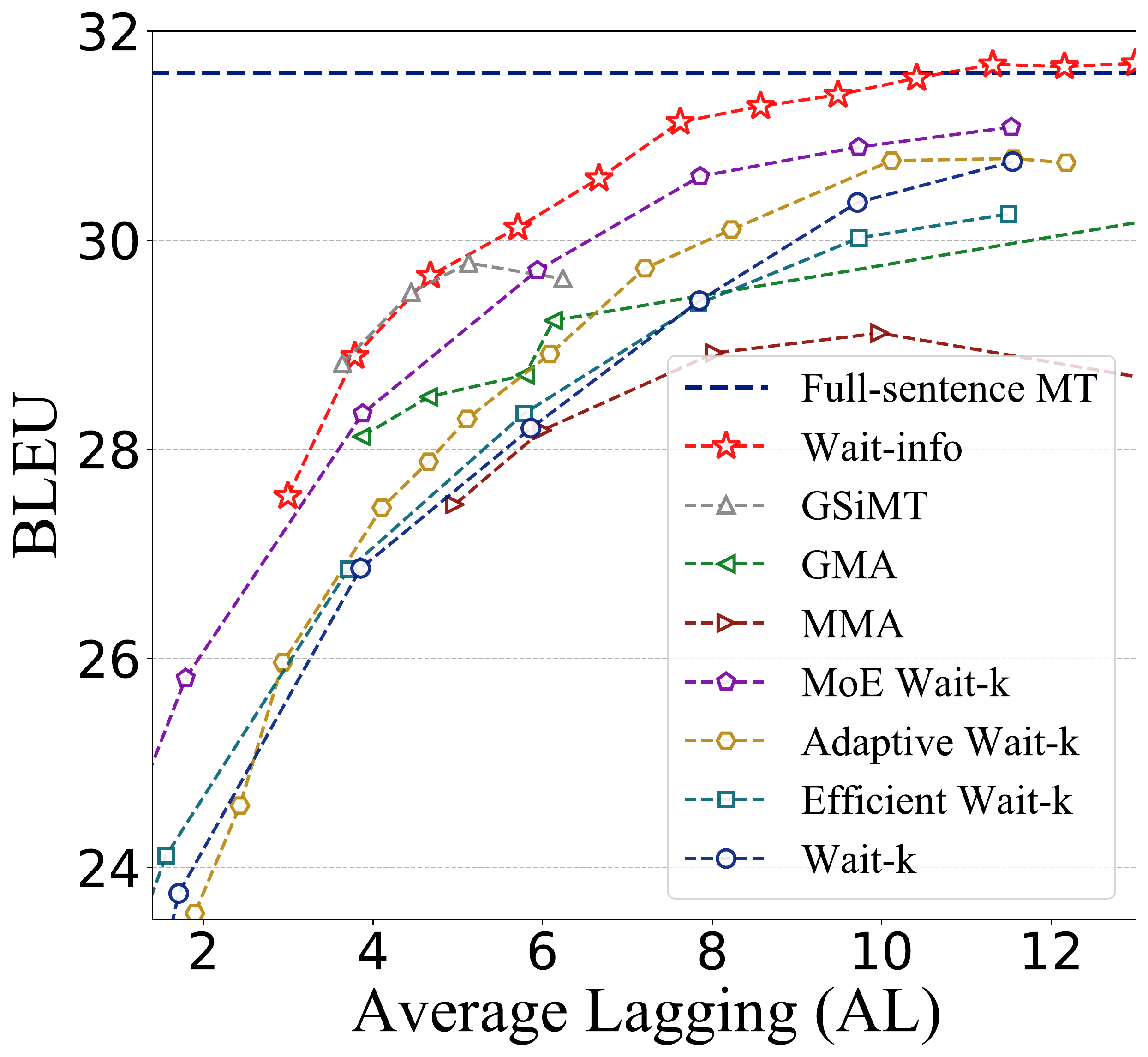}
}
\subfigure[De$\rightarrow$En, Transformer-Big]{
\includegraphics[width=2in]{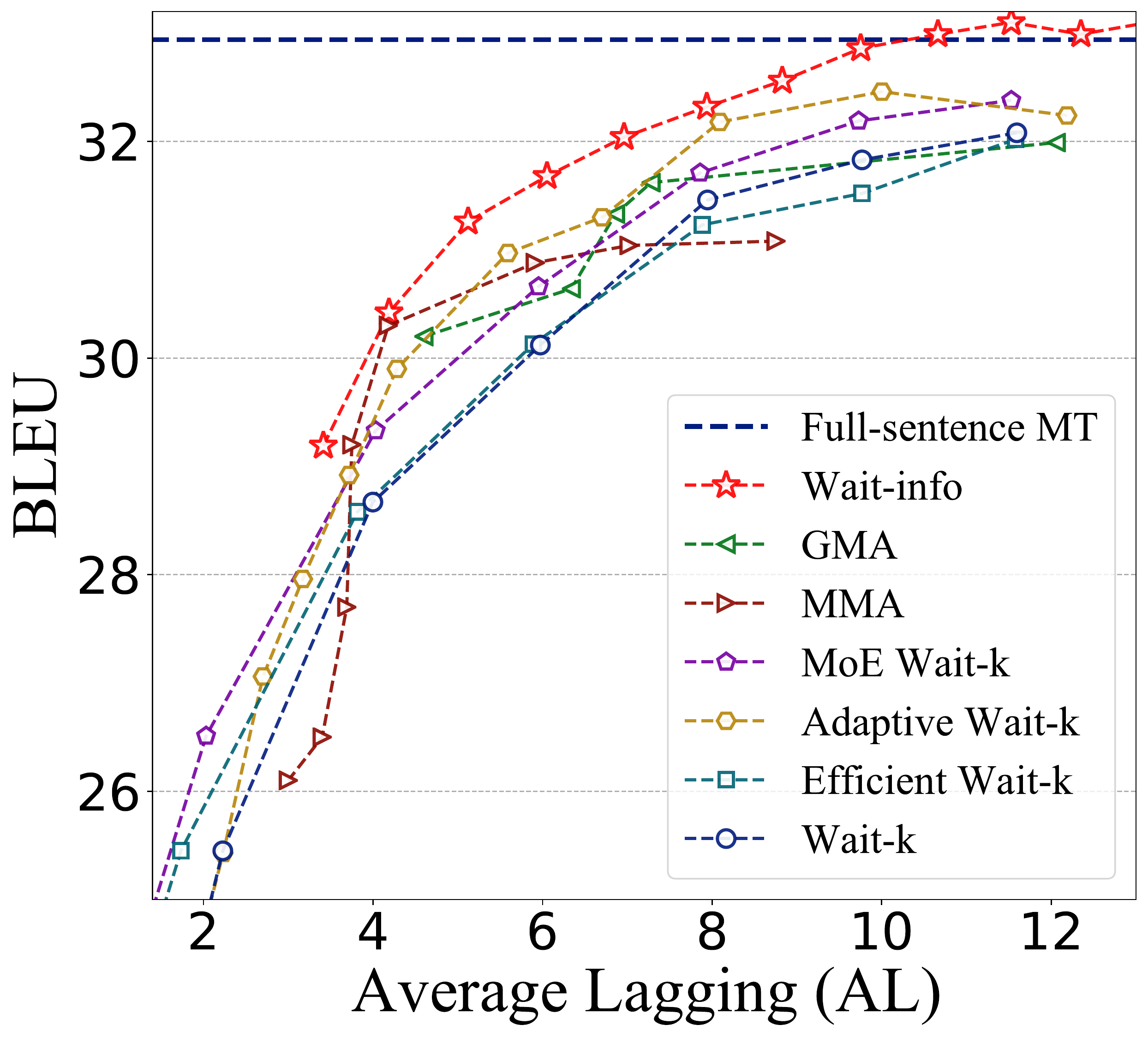}
}
\caption{Translation quality (BLEU) v.s. latency (Average Lagging, AL) of Wait-info and previous methods.}
\label{main}
\end{figure*}

We compare the proposed wait-info policy with previous policies in Figure \ref{main}, where Wait-info outperforms the previous methods under all latency. 
Compared with Wait-k and Efficient Wait-k which directly wait for a fixed number of source tokens, Wait-info balances target outputs and source inputs at the information level, which provides a more flexibly SiMT trade-off and thereby brings significant improvements. MoE Wait-k uses multiple experts to fuse the translation under multiple latency to cope with complex inputs, while Wait-info dynamically adjusts READ/WRITE based on the info and thereby deals with the complex inputs in a more straightforward manner. Both Adaptive Wait-k and Wait-info are adaptive policies, but Adaptive Wait-k still decides which $k$ to use based on the token number of target outputs and received source inputs \cite{zheng-etal-2020-simultaneous}, while Wait-info decides READ/WRITE based on more refined info and thus performs better. Besides, Adaptive Wait-k trains multiple wait-k models, which is computationally expensive, while Wait-info only trains one model to perform SiMT under different latency.

Compared with the adaptive policies, Wait-info also achieves better performance. Previous adaptive policies often decide READ/WRITE based on the current source and target token \cite{Ma2019a,gma}, while Wait-info is based on the accumulated source and target info, which is more reasonable for the SiMT policy. More importantly, most adaptive policies rely on complicated and time-consuming training \cite{zheng-etal-2020-simultaneous} since involving dynamic programming \cite{Ma2019a,miao-etal-2021-generative}. The training of Wait-info is simple as fixed policy, meanwhile the performance is better than adaptive policies.

\begin{figure}[t]
\centering
\subfigure[Effects of two attention.]{
\includegraphics[width=1.46in]{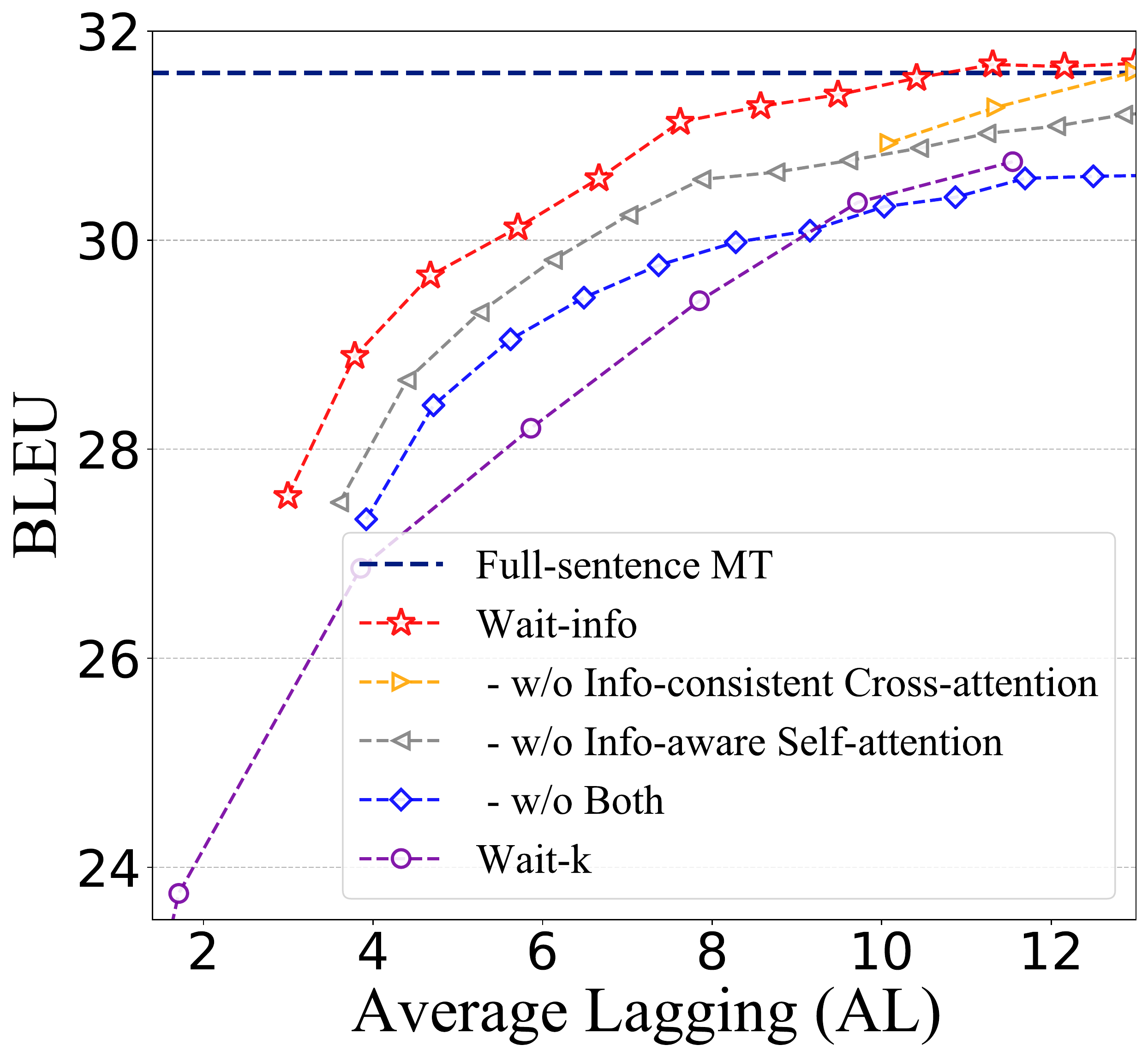} \label{ablation_a}
}\hspace{-2mm}
\subfigure[Effects of src and tgt info.]{
\includegraphics[width=1.46in]{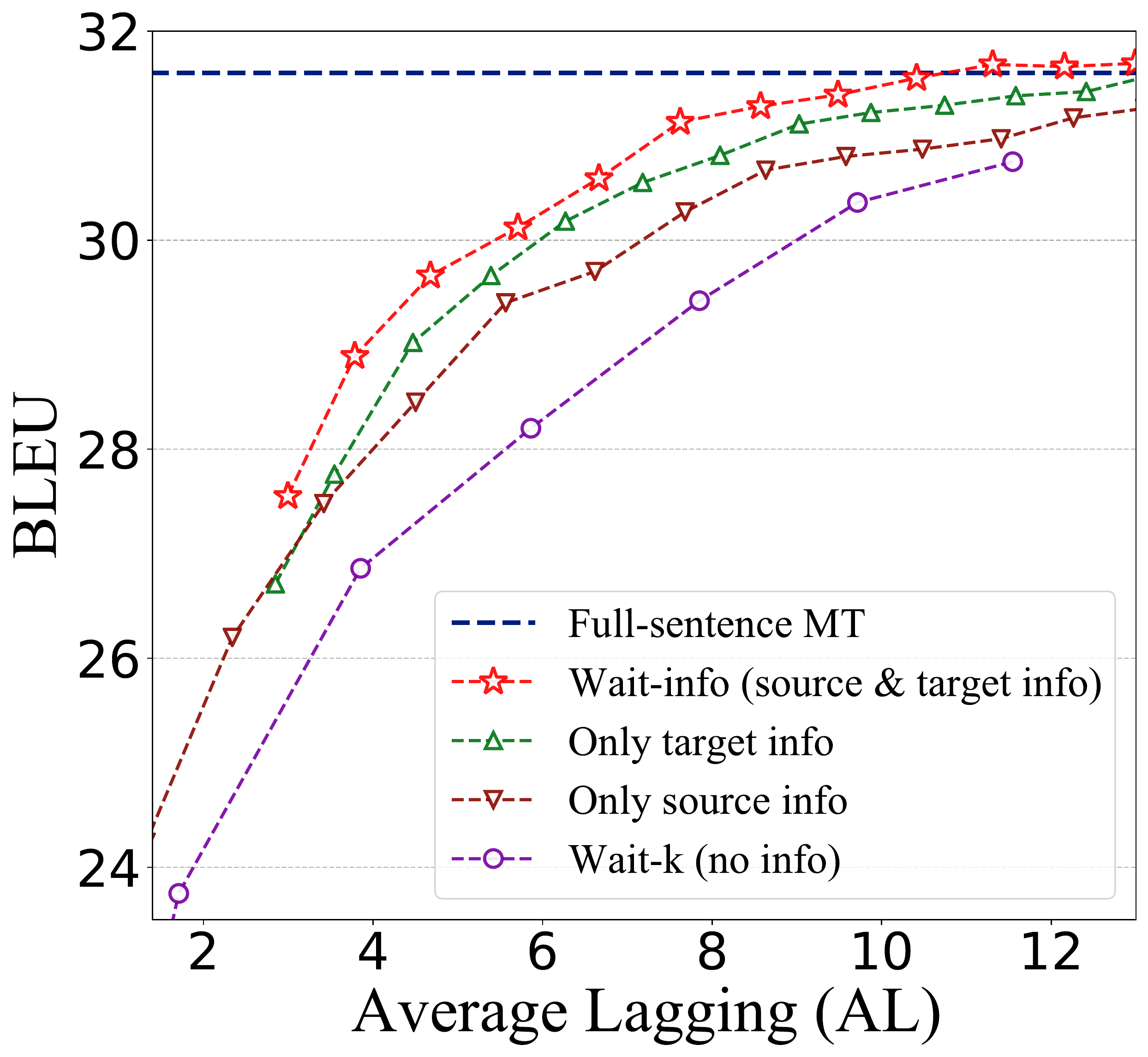} \label{ablation_b}
}
\caption{Ablation Studies on wait-info policy. }
\label{ablation}
\end{figure}

\section{Analysis}
We conduct extensive analyses on wait-info policy. Unless otherwise specified, all results are reported on De$\rightarrow$En with Transformer-Base.

\subsection{Ablation Study}

\textbf{Info-aware Self-attention v.s. Info-consistent Cross-attention}\quad We propose two novel attention to learn the quantified info, so we analyze their roles in Figure \ref{ablation_a}. Without info-aware self-attention, the SiMT performance drops 0.7 BLEU on average, showing that info-aware self-attention is beneficial to the learning of quantified info. When removing the info-consistent cross-attention, the latency becomes much higher, which is because some target info exceptionally becomes much larger than the source info. Info-consistent cross-attention ensures the info consistency between similar tokens and thus controls the latency in a suitable range. When removing both of them, the source or target info is unconstrained and becomes the same value. While the target info will be slightly larger than source info (due to $\mathcal{L}_{sum}$), which is beneficial for SiMT under low latency, we will analyze it in Sec.\ref{sec:length}.

\textbf{Source Info v.s. Target Info}\quad Wait-info policy quantifies the info of both source and target tokens, and we respectively fix the source info $\mathbf{I}^{src}\!=\!\mathbf{1}$ or the target info $\mathbf{I}^{tgt}\!=\!\mathbf{1}$ (i.e., degenerate into wait-k policy that treats each source or target token equally) to compare the effect of only quantifying the source or target info. As shown in Figure \ref{ablation_b}, quantifying the source or target info can both bring significant improvements, where the improvements brought by target info are even more significant.

\subsection{Improvements on Full-sentence MT}
Besides focusing on SiMT, the proposed info-aware Transformer can also improve full-sentence MT. As the full-sentence MT results shown in Table \ref{full}, info-aware Transformer improves 0.08 BLEU on En$\rightarrow$Vi(Small), 0.59 BLEU on De$\rightarrow$En(Base) and 0.39 BLEU on De$\rightarrow$En(Big), showing that explicitly modeling token info is also beneficial for NMT.

\begin{figure*}[t]
\centering
\subfigure[Distribution of source info on different POS.]{
\includegraphics[width=3in]{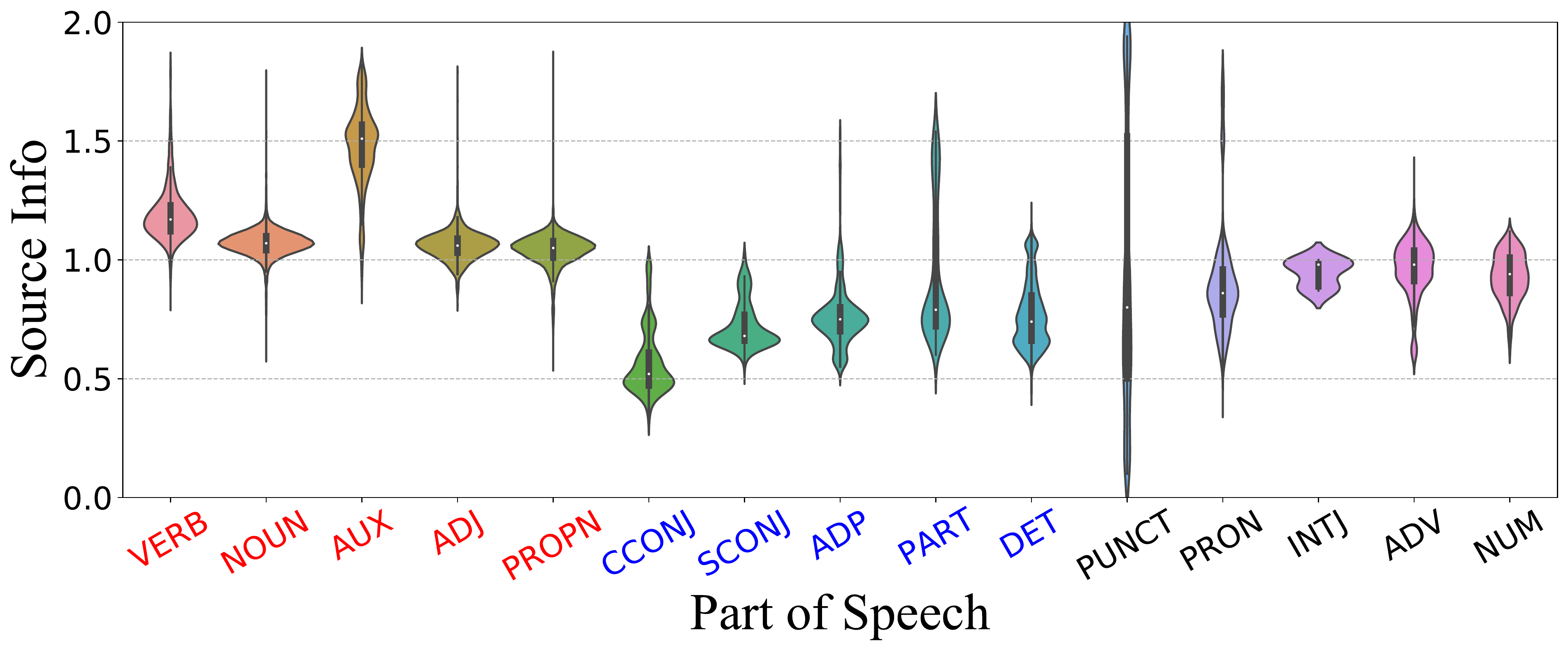} \label{src_pos}
}
\subfigure[Distribution of target info on different POS.]{
\includegraphics[width=3in]{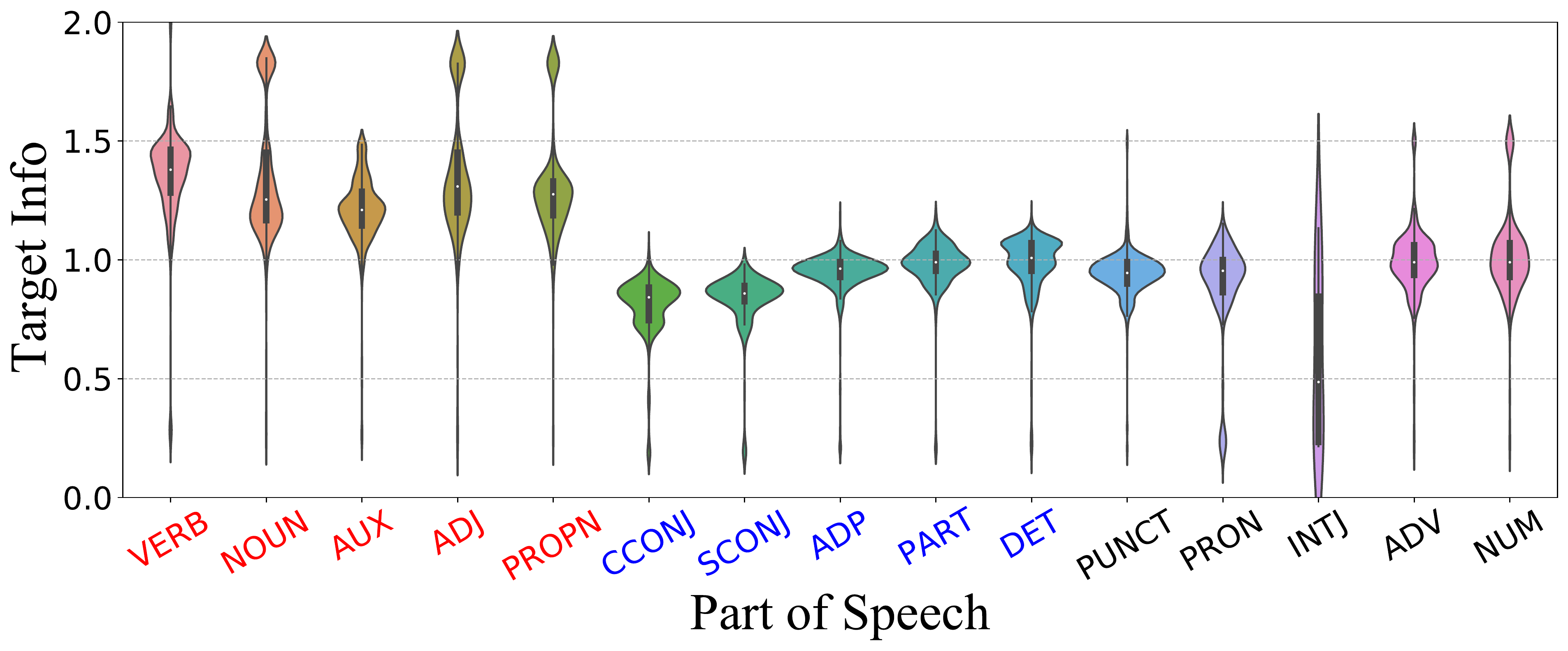} \label{tgt_pos}
}
\caption{Info distribution on different parts of speech (POS), where POS marked in red is often the content word, POS marked in blue is often the function word.}
\label{pos_info}
\end{figure*}

\begin{table}[t]
\centering
\begin{tabular}{l|ccc} \hline
& \textbf{\begin{tabular}[c]{@{}c@{}}En$\rightarrow$Vi\\ (Small)\end{tabular}} & \textbf{\begin{tabular}[c]{@{}c@{}}De$\rightarrow$En\\ (Base)\end{tabular}} &  \textbf{\begin{tabular}[c]{@{}c@{}}De$\rightarrow$En\\ (Big)\end{tabular}} \\ \hline
Transformer                                                      & 28.90          & 31.60                                                          & 32.84                                                          \\ \hline
\begin{tabular}[c]{@{}l@{}}Info-aware\\ Transformer\end{tabular} & \textbf{28.98} & \textbf{32.19}                                                  & \textbf{33.23}      \\\hline                                           
\end{tabular}
\caption{Improvements on full-sentence MT.}
\label{full}
\end{table}

\subsection{Comparison on Information Modeling}

\begin{figure}[t]
\centering
\includegraphics[width=2.3in]{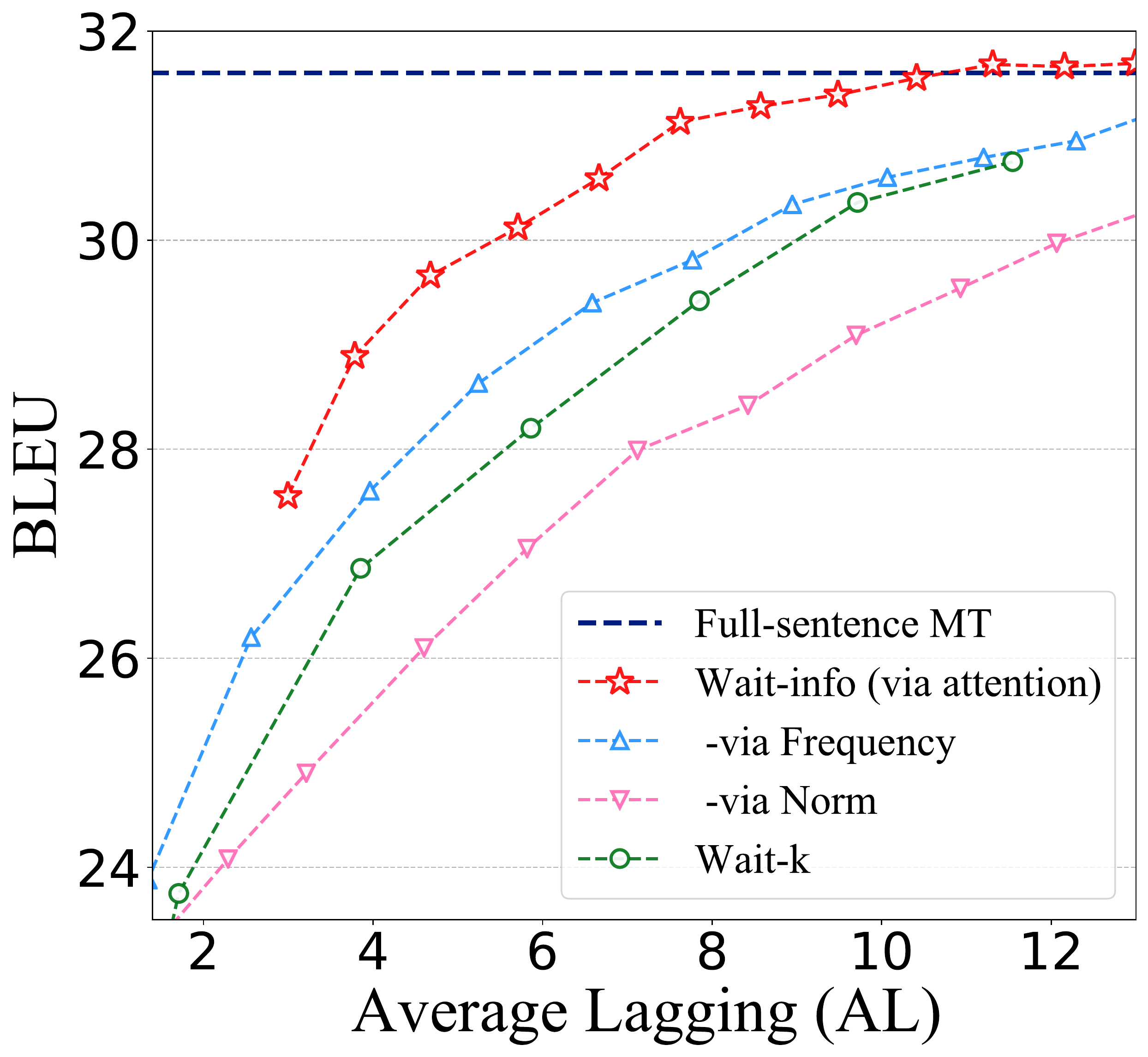}
\caption{Comparison of different methods of information modeling in wait-info policy, including via attention, token frequency and embedding norm.}
\label{ab_freq_norm}
\end{figure}

To model the information amount contained in each token, we propose an unsupervised method to adaptively learn the info from the attention mechanism. Some previous methods apply heuristic methods to model the information, such as using the token frequency to indicate the amount of information \cite{moradi-etal-2019-interrogating,chen-etal-2020-content} or associating the norm of embedding with the token information \cite{liu-etal-2020-norm,kobayashi-etal-2020-attention}. We apply different methods of information modeling (i.e., via attention, via token frequency and via norm of token embedding) in the proposed wait-info policy, and show the results in Figure \ref{ab_freq_norm}.

Using embedding norm to indicate token info is not suitable for the proposed wait-info policy, we argue that this is because the embedding norm is better at identifying specific tokens such as <eos> and punctuation \cite{kobayashi-etal-2020-attention}, but has limited ability to distinguish token information in more detail. Modeling the info via attention and frequency can both achieve improvements, where our proposed method of learning info from attention performs much better, since jointly learning the info with translation is more flexible than the fixed frequency \cite{zhang-etal-2022-conditional}.

\subsection{Quality of Quantified Info}

We expect that the proposed info can reflect the amount of information contained in the token, thus providing reasonable evidence for the SiMT policy. To verify the quality of quantified info, we further explore whether the quantified info can distinguish different types of tokens, especially content words and function words as mentioned above. In response to this question, we categorize different tokens using the Universal Part-of-Speech (POS) Tagging tool\footnote{\url{huggingface.co/flair/upos-multi}}, and draw the info distribution of tokens with different POS\footnote{VERB: verb, NOUN: noun, AUX: auxiliary, ADJ: adjective, PROPN: proper noun, CCONJ: coordinating conjunction, SCONJ: subordinating conjunction, ADP: adposition, PART: particle, DET: determiner, PUNCT: punctuation, PRON: pronoun, INTJ: interjection, ADV: adverb, NUM: numeral.} via violin plot in Figure \ref{pos_info}. Tokens with different parts of speech have obvious differences in info distribution, where content words (e.g., VERB, NOUN, AUX, ADJ, PROPN) generally get larger info, while function words (e.g., CCONJ, SCONJ, ADP, PART, DET) have smaller info, which is in line with our expectations \cite{xu-etal-2019-leveraging}. Therefore, info can successfully learn the amount of information contained in different tokens, so as to develop a reasonable SiMT policy.

\subsection{Flexibility on Length Difference}
\label{sec:length}

\begin{table}[t]
\centering
\begin{tabular}{l|ccc|c} \hline
        & \multicolumn{3}{c|}{Length Ratio (src/tgt)} & \multirow{2}{*}{\begin{tabular}[c]{@{}c@{}}Info Ratio\\ (tgt/src)\end{tabular}} \\
        & Train.        & Dev.        & Test.        &                                                                                 \\ \hline
En$\rightarrow$Vi & 0.84          & 0.84        & 0.81         & 0.85                                  \\
De$\rightarrow$En   & 1.09          & 1.08        & 1.06         & 1.10    \\\hline                             
\end{tabular}
\caption{Length ratio (source/target) on En$\rightarrow$Vi and De$\rightarrow$En and the info ratio (target/source) in our wait-info policy. During training, the ratio between source and target info is successfully adjusted according to the length ratio, thereby ensuring that the total source info and total target info are equal.}
\label{data_rate}
\end{table}

\textbf{Early-stop Caused by Length Difference}\quad The length difference between the two languages is a major challenge for SiMT, especially for wait-k policy. Wait-k policy is sensitive to the length ratio between source and target and sometimes may force the model to finish the target translation before reading the complete source sentence \cite{ma-etal-2019-stacl,laf}, named \emph{early-stop}, especially when the source sentence is longer than the target sentence. Formally, wait-k policy will early-stop translating  when $g_{k}(m)<n$, where $g_{k}(m)\!=\!k\!+\!m\!-\!1$ defined in Eq.(\ref{eq3}), $n$ and $m$ are source and target lengths. 

More importantly, the length difference is always language-specific \cite{ma-etal-2019-stacl}, and Table \ref{data_rate} reports the length ratio between source and target on En$\rightarrow$Vi and De$\rightarrow$En datasets. As seen, the target sentence in En$\rightarrow$Vi is generally longer than the source sentence, on the contrary, the source sentence in De$\rightarrow$En is longer (i.e., $n\!>\!m$), which is more prone to the early-stop. To study the severity of early-stop, we calculate the proportion of early-stop in wait-k policy in Table 2, where over 20\% of De$\rightarrow$En cases will early stop translating before receiving the complete source sentence under low latency. The essential reason for early-stop is that wait-k policy balances source and target at the token level, where the token-level balance is not the best choice because the number of tokens (i.e., length) is often language-specific.

\begin{table}[t]
\centering
\begin{tabular}{ccc|cc} \hline
\multicolumn{3}{c|}{\textbf{Wait-k}} & \multicolumn{2}{c}{\textbf{Wait-info}} \\\hline
$k$      & De$\rightarrow$En        & En$\rightarrow$Vi       & De$\rightarrow$En              & En$\rightarrow$Vi             \\\hline
1      & 29.88\%      & 0.39\%      & 0.00\%             & 0.00\%            \\
3      & 22.68\%      & 0.16\%      & 0.00\%             & 0.00\%            \\
5      & 13.09\%      & 0.00\%      & 0.00\%             & 0.00\%            \\
7      & $\;\;$6.78\%       & 0.00\%      & 0.00\%             & 0.00\%            \\
9      & $\;\;$3.23\%       & 0.00\%      & 0.00\%             & 0.00\%           \\\hline
\end{tabular}
\caption{Proportion of early-stop. Under low latency, Wait-k emerges much early-stop on De$\rightarrow$En, while Wait-info completely avoids this situation (0.00\%). Note that for Wait-info, we select the results under the similar latency with the Wait-k for comparison.}
\label{early_stop}
\end{table}
\begin{figure}[t]
\centering
\includegraphics[width=2.3in]{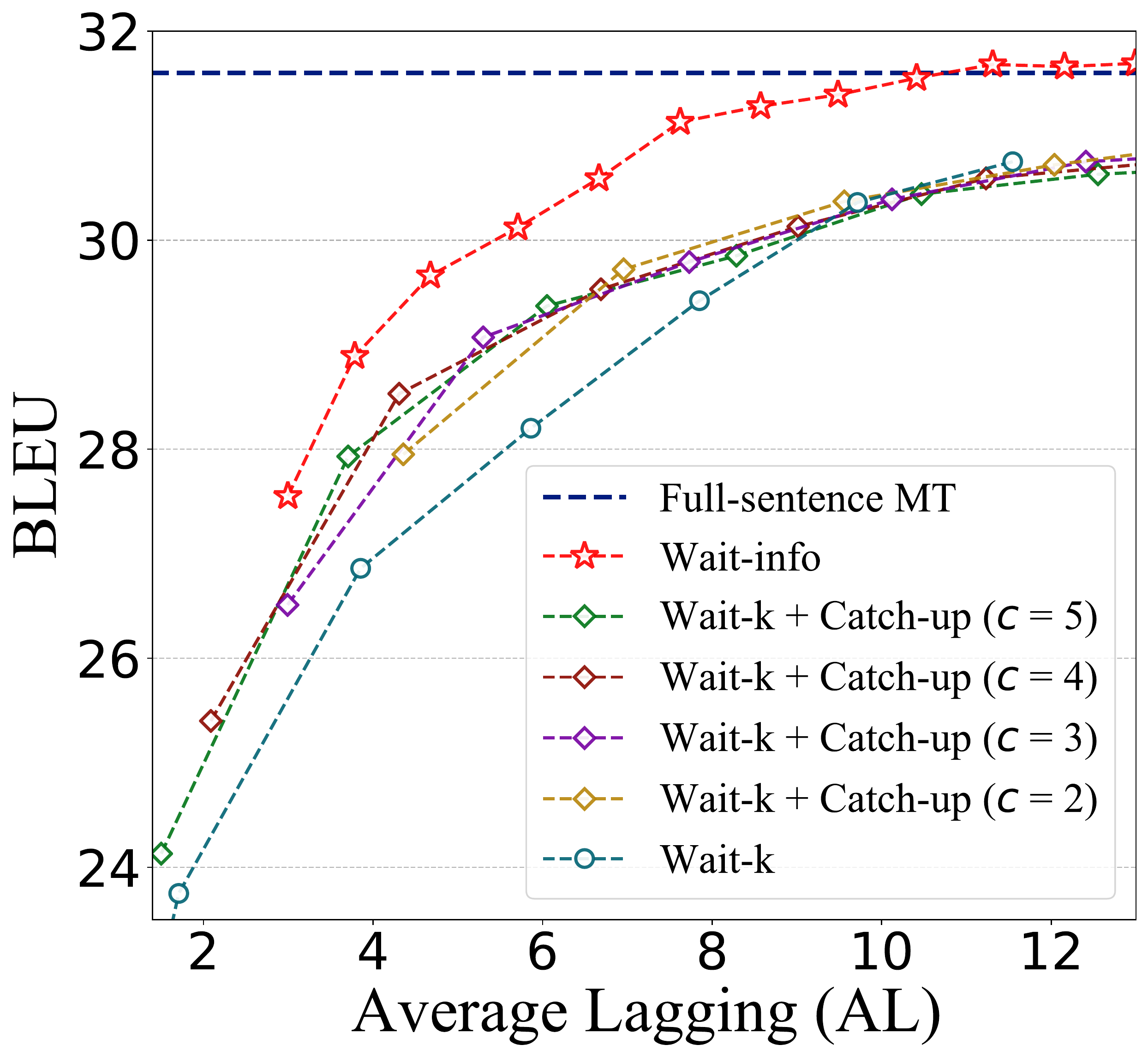}
\caption{Comparison of Wait-info and Catch-up.}
\label{ab_catchup}
\end{figure}

\textbf{Wait-info Avoids Early-stop}\quad Owing to $\mathcal{L}_{sum}$ in Eq.(\ref{eq6}) that constrains the total source info to be equal to total target info, the proposed wait-info policy can learn to adjust the ratio between source and target info according to the length ratio, thereby avoiding early-stop. As shown in Table \ref{data_rate}, the average quantified info ratio (target info/source info) is basically the same as the length ratio (source length/target length), which shows that $\mathcal{L}_{sum}$ successfully constrains the equality between total source info and total target info. Therefore, as shown in Table \ref{early_stop}, wait-info policy completely avoids the early-stop caused by length difference. Different from the wait-k policy, wait-info policy balances source and target at the info level, where the total info of target and source is the same and language-independent, thereby overcoming the length difference between two languages.

\begin{figure*}[t]
\centering
\includegraphics[width=6.31in]{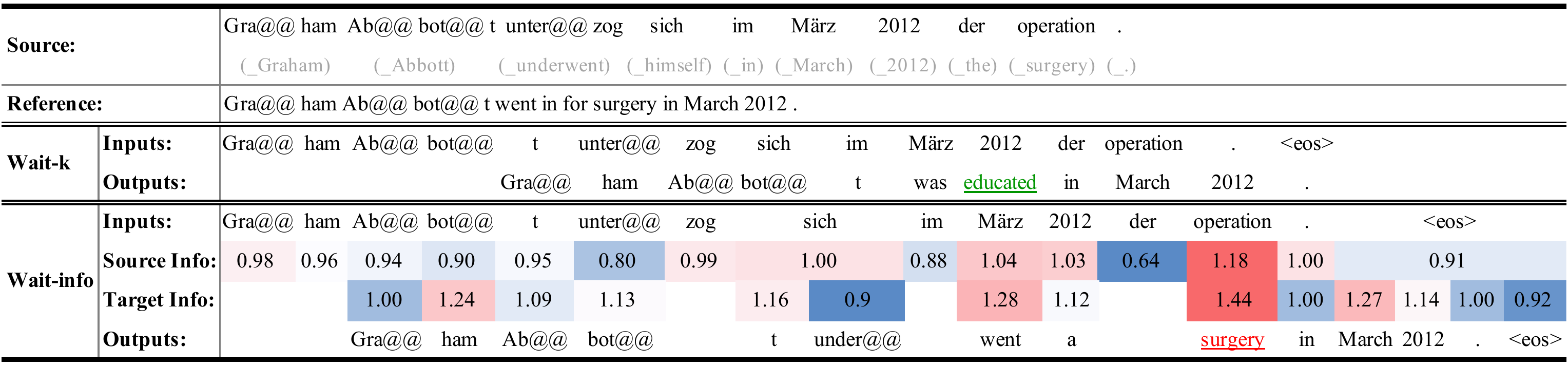}
\caption{Case study of No.1219 in De$\rightarrow$En test set, showing Wait-k ($k=5$) and Wait-info ($\mathcal{K}=1$) under the similar latency (AL $\approx 3$). To show the process of SiMT more clearly, we correspond the outputs and inputs in the horizontal direction, indicating which source tokens are received when translating the target token. For source and target info, values that are larger than the average info (i.e., containing more information) are marked in red, values that are smaller than the average info (i.e., containing less information) are marked in blue.}
\label{case1219}
\end{figure*}

\textbf{Wait-info v.s. Catch-up}\quad
To avoid early-stop, \citet{ma-etal-2019-stacl} proposed a heuristic approach \emph{Catch-up} for wait-k policy to compensate for the length difference between target and source. Catch-up requires the model to read one additional source token after every generating $c$ target tokens (i.e., try to read more source tokens to avoid early-stop), where $c$ is a hyperparameter. We compare the performance of `Wait-k+Catch-up' and Wait-info in Figure \ref{ab_catchup}, where Wait-info performs better since it balances the source and target more flexibly from the info level rather than reading more source tokens according to heuristic rules.

\section{Case Study}
\label{sec:appendix}

To study the specific improvement of the proposed wait-info policy compared to the wait-k policy, we conduct a case study in Figure \ref{case1219}. In Wait-k, the model is forced to wait for a fixed 5 tokens before translating, which makes the model either too aggressive or too conservative in different cases \cite{zheng-etal-2020-simultaneous}. As shown in this case, at the beginning of translation, when translating `\textit{Grahams}', 2 source tokens are enough to translate, but wait-k policy forces the model to wait for 5 tokens, resulting in unnecessary waiting. When translating the noun `\textit{surgery}', the model should have waited until receiving `\textit{operation}', but the model was forced to output in advance, resulting in the wrong translation `\textit{educated}' (marked in green).

In Wait-info, this weakness is ameliorated by quantifying the information in each token rather than considering each token equally. First of all, we find the proposed info can effectively distinguish different tokens, where the content words often get larger info, such as `\textit{sich}', `\textit{März}' and `\textit{operation}' in German, and `\textit{went}', `\textit{surgery}' and `\textit{March}' in English, thereby being more important to the SiMT policy. Owing to the quantified info, when translating the `\textit{surgery}', the model recognized that the previous `\textit{der}' (i.e., determiner in German) does not contain enough info, so the model continues to wait for the `\textit{operation}' and thereby generates the correct translation `\textit{surgery}' (marked in red). Overall, in wait-info policy, tokens with larger info, such as verbs and nouns, play a more important role in the model's decision of READ/WRITE, making it easier to ensure that those content words are read before translating.

\section{Conclusion}
In this paper, we quantify the information in tokens and propose a wait-info policy accordingly. Experiments show the superiority of our method on SiMT tasks and good explainability of the quantified info.

\section*{Limitations}
In this work, we quantify the amount of information contained in each token via a scalar. Although quantifying information as a scalar is intuitive and friendly to SiMT policy, the expression space of a scalar may be limited for some particularly complex situations. Quantifying the information contained in each token through a low-dimensional vector may be able to further improve the performance of wait-info policy. However, how to balance the info in vector form between source and target is also a new challenge, and we will put it into our future work.

\section*{Acknowledgements}
We thank all the anonymous reviewers for their insightful and valuable comments.

\bibliography{anthology,custom}
\bibliographystyle{acl_natbib}

\newpage

\appendix

\section{Comparison on Settings of Total Info}
\begin{figure}[t]
\centering
\includegraphics[width=2.8in]{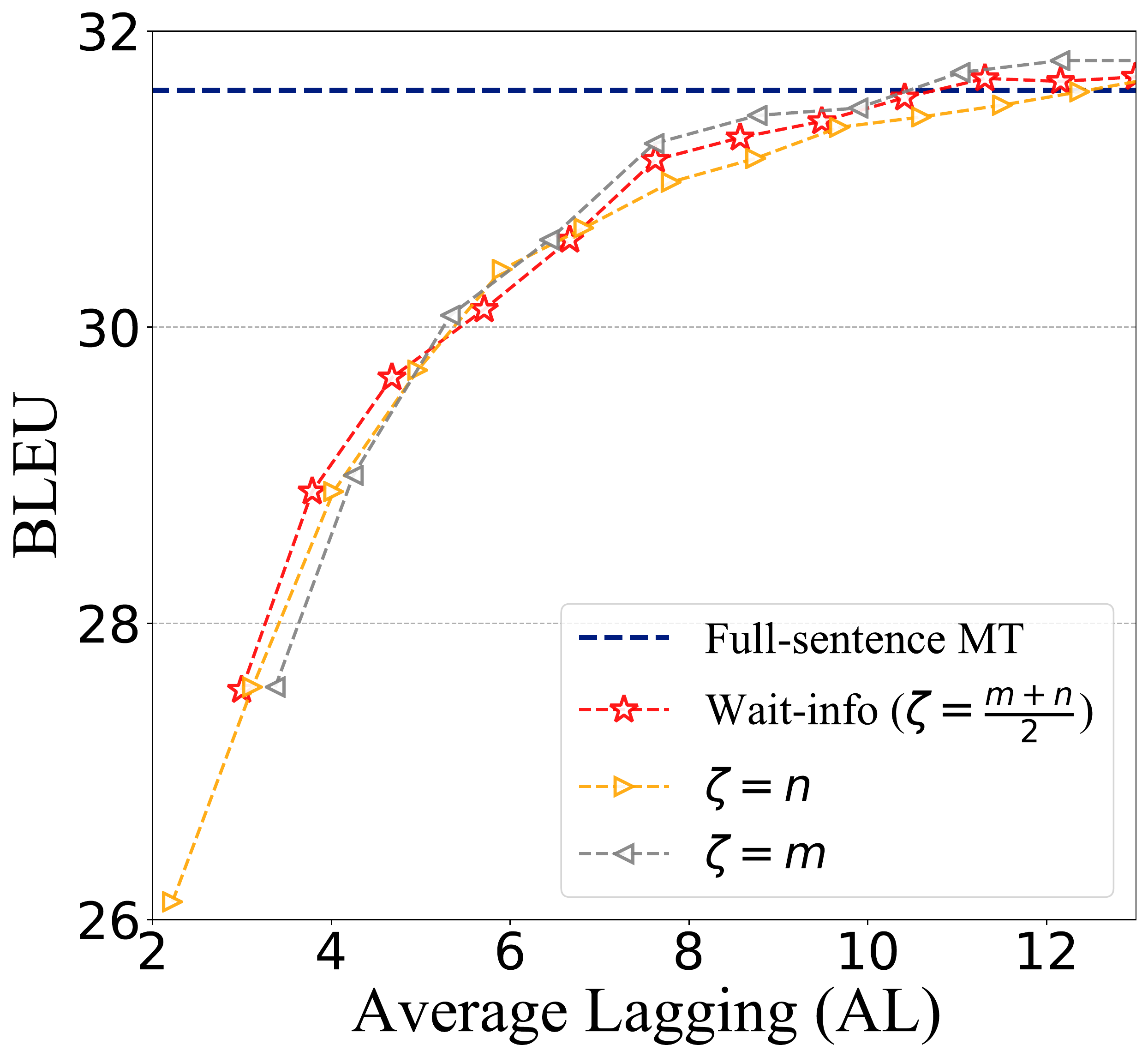}
\caption{Comparison on different settings of total info $\zeta$ in Eq.(\ref{eq6}), where $n$ is the length of source sentence and $m$ is the length of source sentence.}
\label{total_info}
\end{figure}

Based on the semantic equivalence between the source sentence and the target sentence, we introduce $\mathcal{L}_{sum}$ to constrain the total info of the source tokens and target tokens in Eq.(\ref{eq6}). $\mathcal{L}_{sum}$ can not only ensure that the total info of the source and target is equal, but also constrain the average info to be around 1, which is friendly to wait-info policy. In our experiments, we set the total info $\zeta=\frac{m+n}{2}$, where $n$ is the length of source sentence and $m$ is the length of source sentence. We compare the performance under different $\zeta$ settings in Figure \ref{total_info}, including $\zeta=\frac{m+n}{2}$, $\zeta=m$ and $\zeta=n$. Our method is not sensitive to the setting of $\zeta$ and achieves almost similar performance under different settings.

\section{Extended Analyses on Early-stop}
\label{app:early-stop}

\begin{figure}[t]
\centering
\includegraphics[width=3in]{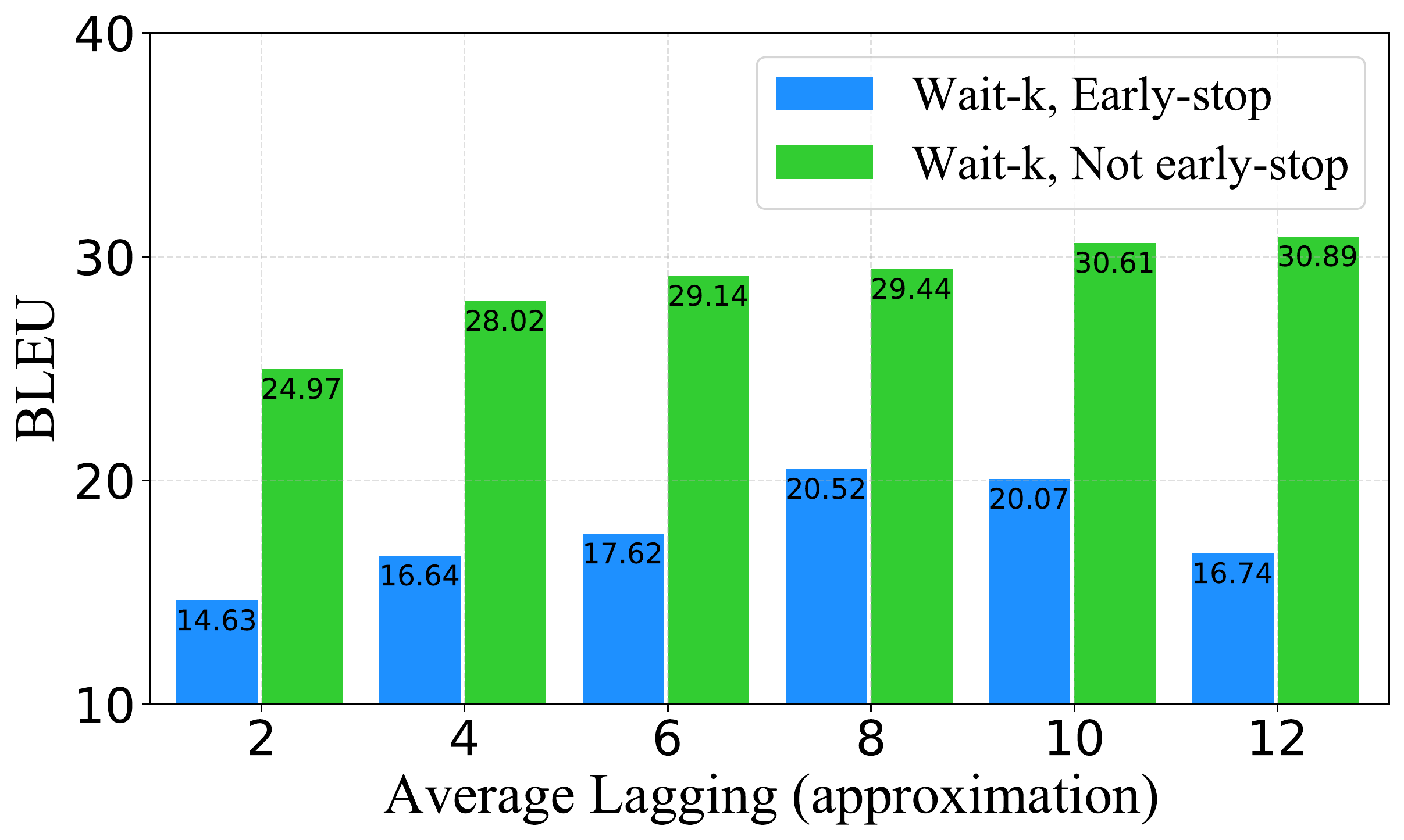}
\caption{We divide the De$\rightarrow$En test set into two sets, \emph{early-stop} and \emph{not-early-stop}, based on whether the wait-k early-stop translating before receiving the complete source inputs. Then we calculate the BLEU scores of wait-k policy on each set. }
\label{earlystopbleu}
\end{figure}

\textbf{Severity of Early-stop} As mentioned in Sec.\ref{sec:length}, wait-k policy may early-stop translating before receiving complete source inputs, especially under low latency. The reason for early-stop is $g_{k}(m)<n$ caused by the length difference between the source and target. To investigate how seriously early-stop affects translation quality, we calculate the BLEU scores of wait-k policy for early-stop or not-early-stop cases respectively in Figure \ref{earlystopbleu}. When the wait-k policy appears early-stop, the translation quality is 11 BLEU lower than those cases not-early-stop on average, indicating that early-stop seriously affects SiMT performance.

\textbf{Why Does Wait-info Avoid Early-stop?} The wait-k policy will early-stop translating when $g_{k}(m)\!<\!n$. While for wait-info policy, $g_{\mathcal{K}}(m)\!=\!\underset{j}{\!\mathrm{argmin}}\!\left (  \sum_{l=1}^{j}\! I^{src}_{l} \geq \sum_{l=1}^{m}\! I^{tgt}_{l}\!+\!\mathcal{K}\right )$ (defined in Eq.(\ref{eq11})) will almost always greater than $n$, since we introduce an info-sum loss $\mathcal{L}_{sum}$ (defined in Eq.(\ref{eq6})) to constrain the $\sum_{j=1}^{n}I^{src}_{j}=\sum_{i=1}^{m}I^{tgt}_{i}$.

\section{Numerical Results}

Besides Average Lagging (AL) \cite{ma-etal-2019-stacl}, we also use Consecutive Wait (CW) \cite{gu-etal-2017-learning}, Average Proportion (AP) \cite{Cho2016} and Differentiable Average Lagging (DAL) \cite{Arivazhagan2019} to evaluate the latency of the SiMT model. We use $g\left(i\right)$ to record the number of source tokens received when translating $y_{i}$. The calculation of latency metrics are as follows.

\textbf{Consecutive Wait (CW)} \cite{gu-etal-2017-learning} evaluates the average number of source tokens waited between two target tokens, calculated as:
\begin{gather}
    \mathrm{CW}=\frac{\sum_{i=1}^{\left | \mathbf{y} \right |} (g\left(i\right)-g\left(i-1\right))}{\sum_{i=1}^{\left | \mathbf{y} \right |}\mathbbm{1}_{g\left(i\right)-g\left(i-1\right)>0}},
\end{gather}
where $\mathbbm{1}_{g\left(i\right)-g\left(i-1\right)}=1$ counts the number of $g\left(i\right)-g\left(i-1\right)>0$.

\textbf{Average Proportion (AP)} \cite{Cho2016} measures the proportion of the received source tokens, calculated as:
\begin{gather}
    \mathrm{AP}=\frac{1}{\left | \mathbf{x} \right | \left | \mathbf{y} \right |}\sum_{i=1}^{\left | \mathbf{y} \right |} g\left(i\right).
\end{gather}

\textbf{Differentiable Average Lagging (DAL)} \cite{Arivazhagan2019} is a differentiable version of average lagging, calculated as:
\begin{align}
g^{'}\left(i\right)=&\;\left\{\begin{matrix}
g\left(i\right) & i=1\\ 
 \mathrm{max}\left (g\left(i\right),g^{'}\left(i-1\right)+ \frac{\left | \mathbf{x} \right |}{\left | \mathbf{y} \right |} \right )& i>1
\end{matrix}\right., \\
    \mathrm{DAL}=&\;\frac{1}{\left | \mathbf{y} \right | }\sum\limits_{i=1}^{\left | \mathbf{y} \right |}g^{'}\left(i\right)-\frac{i-1}{\left | \mathbf{x} \right |/\left | \mathbf{y} \right |}.
\end{align}

\textbf{Numerical Results}\quad Table \ref{res_envi_small}, \ref{res_deen_base} and \ref{res_deen_big} report the numerical results of all systems in our experiments, evaluated with BLEU for translation quality and CW, AP, AL and DAL for latency.

\begin{table*}[b]
\centering
\scriptsize
\begin{tabular}{l|cccccc} \hlinew{1.2pt}
\multicolumn{7}{c}{\textbf{IWSLT15 English$\rightarrow$Vietnamese$\;\;\;\;\;$Transformer-Small}}                                                             \\  \hline
\multirow{2}{*}{\begin{tabular}[c]{@{}l@{}}\textbf{Full-sentence} MT\\\cite{NIPS2017_7181}\end{tabular}} &            & CW    & AP   & AL    & DAL   & BLEU  \\
                                           &            & 22.08 & 1.00 & 22.08 & 22.08 & 28.91 \\ \hline
\multirow{6}{*}{\begin{tabular}[c]{@{}l@{}}\textbf{Wait-k}\\\cite{ma-etal-2019-stacl}\end{tabular}}           & $k$          & CW    & AP   & AL    & DAL   & BLEU  \\
                                           & 1          & 1.00  & 0.63 & 3.03  & 3.54  & 25.21 \\
                                           & 3          & 1.17  & 0.71 & 4.80  & 5.42  & 27.65 \\
                                           & 5          & 1.46  & 0.78 & 6.46  & 7.06  & 28.34 \\
                                           & 7          & 1.96  & 0.83 & 8.21  & 8.79  & 28.60 \\
                                           & 9          & 2.73  & 0.88 & 9.92  & 10.51 & 28.69 \\ \hline
\multirow{6}{*}{\begin{tabular}[c]{@{}l@{}}\textbf{Efficient Wait-k}\\\cite{multipath}\end{tabular}} & $k$          & CW    & AP   & AL    & DAL   & BLEU  \\
                                           & 1          & 1.01  & 0.63 & 3.06  & 3.61  & 26.23 \\
                                           & 3          & 1.17  & 0.71 & 4.66  & 5.20  & 28.21 \\
                                           & 5          & 1.46  & 0.78 & 6.38  & 6.94  & 28.56 \\
                                           & 7          & 1.96  & 1.96 & 8.13  & 8.69  & 28.62 \\
                                           & 9          & 2.73  & 0.87 & 9.80  & 10.34 & 28.52 \\ \hline
\multirow{8}{*}{\begin{tabular}[c]{@{}l@{}}\textbf{Adaptive Wait-k}\\\cite{zhang-etal-2020-learning-adaptive}\end{tabular}}  & ( $\rho_{1}$, $\rho_{9}$ )   & CW    & AP   & AL    & DAL   & BLEU  \\
                                           & (0.02, 0.00)   & 1.05  & 0.63 & 2.98  & 3.64  & 25.69 \\
                                           & (0.04, 0.00)   & 1.19  & 0.63 & 3.07  & 4.06  & 26.05 \\
                                           & (0.05, 0.00)   & 1.27  & 1.27 & 3.14  & 4.30  & 26.33 \\
                                           & (0.10, 0.00)    & 1.97  & 0.68 & 4.08  & 6.05  & 27.80 \\
                                           & (0.10, 0.05) & 2.36  & 0.71 & 4.77  & 7.11  & 28.46 \\
                                           & (0.20, 0.00)    & 2.73  & 0.78 & 6.56  & 8.34  & 28.73 \\
                                           & (0.30, 0.20)  & 3.39  & 0.86 & 9.42  & 10.42 & 28.80 \\ \hline
\multirow{6}{*}{\begin{tabular}[c]{@{}l@{}}\textbf{MoE Wait-k}\\\cite{zhang-feng-2021-universal}\end{tabular}}       & $k$          & CW    & AP   & AL    & DAL   & BLEU  \\
                                           & 1          & 1.00  & 0.63 & 3.19  & 3.76  & 26.56 \\
                                           & 3          & 1.17  & 0.71 & 4.70  & 5.42  & 28.43 \\
                                           & 5          & 1.46  & 0.78 & 6.43  & 7.14  & 28.73 \\
                                           & 7          & 1.97  & 0.83 & 8.19  & 8.88  & 28.81 \\
                                           & 9          & 2.73  & 0.87 & 9.86  & 10.39 & 28.88 \\ \hline
\multirow{7}{*}{\begin{tabular}[c]{@{}l@{}}\textbf{MMA}\\\cite{Ma2019a}\end{tabular}}              & $\lambda$     & CW    & AP   & AL    & DAL   & BLEU  \\
                                           & 0.4        & 1.03  & 0.58 & 2.68  & 3.46  & 27.73 \\
                                           & 0.3        & 1.09  & 0.59 & 2.98  & 3.81  & 27.90 \\
                                           & 0.2        & 1.15  & 0.63 & 3.57  & 4.44  & 28.47 \\
                                           & 0.1        & 1.31  & 0.67 & 4.63  & 5.65  & 28.42 \\
                                           & 0.04       & 1.64  & 0.70 & 5.44  & 6.57  & 28.33 \\
                                           & 0.02       & 2.01  & 0.76 & 7.09  & 8.29  & 28.28 \\ \hline
\multirow{6}{*}{\begin{tabular}[c]{@{}l@{}}\textbf{GMA}\\\cite{gma}\end{tabular}}              & $\delta$     & CW    & AP   & AL    & DAL   & BLEU  \\
                                           & 0.9        & 1.20  & 0.65 & 3.05  & 4.08  & 27.95 \\
                                           & 1.0        & 1.27  & 0.68 & 4.01  & 4.77  & 28.20 \\
                                           & 2.0        & 1.49  & 0.74 & 5.47  & 6.37  & 28.44 \\
                                           & 2.2        & 1.60  & 0.77 & 6.04  & 6.96  & 28.56 \\
                                           & 2.5        & 1.74  & 0.78 & 6.55  & 7.55  & 28.72 \\ \hline
\multirow{9}{*}{\textbf{Wait-info}}        & $\mathcal{K}$          & CW    & AP   & AL    & DAL   & BLEU  \\
                                           & 1          & 1.10  & 0.67 & 3.76  & 4.33  & 28.37 \\
                                           & 2          & 1.19  & 0.69 & 4.10  & 4.71  & 28.45 \\
                                           & 3          & 1.34  & 0.71 & 4.60  & 5.28  & 28.54 \\
                                           & 4          & 1.46  & 0.74 & 5.28  & 5.97  & 28.59 \\
                                           & 5          & 1.63  & 0.77 & 6.01  & 6.71  & 28.70 \\
                                           & 6          & 1.86  & 0.80 & 6.80  & 7.51  & 28.78 \\
                                           & 7          & 2.16  & 0.82 & 7.61  & 8.33  & 28.80 \\
                                           & 8          & 2.51  & 0.84 & 8.39  & 9.11  & 28.82 \\ \hlinew{1.2pt}
\end{tabular}
\caption{Numerical results on En$\rightarrow$Vi with Transformer-Small.}
\label{res_envi_small}
\end{table*}

\begin{table*}[t]
\centering
\scriptsize
\begin{tabular}{l|cccccc}  \hlinew{1.2pt}
\multicolumn{7}{c}{\textbf{WMT15 German$\rightarrow$English$\;\;\;\;\;$Transformer-Base}}                                                      \\ \hline
\multirow{2}{*}{\begin{tabular}[c]{@{}l@{}}\textbf{Full-sentence} MT\\\cite{NIPS2017_7181}\end{tabular}} &            & CW    & AP   & AL    & DAL   & BLEU \\
                                           &            & 27.77 & 1.00 & 27.77 & 27.77 & 31.60   \\ \hline
\multirow{8}{*}{\begin{tabular}[c]{@{}l@{}}\textbf{Wait-k}\\\cite{ma-etal-2019-stacl}\end{tabular}}           & $k$          & CW    & AP   & AL    & DAL   & BLEU    \\
                                           & 1          & 1.17  & 0.52 & 0.02  & 1.84  & 17.61   \\
                                           & 3          & 1.23  & 0.59 & 1.71  & 3.33  & 23.75   \\
                                           & 5          & 1.37  & 0.66 & 3.85  & 5.20  & 26.86   \\
                                           & 7          & 1.70  & 0.73 & 5.86  & 7.12  & 28.20   \\
                                           & 9          & 2.17  & 0.78 & 7.85  & 9.01  & 29.42   \\
                                           & 11         & 2.78  & 0.82 & 9.71  & 10.79 & 30.36   \\
                                           & 13         & 3.56  & 0.86 & 11.55 & 12.49 & 30.75   \\ \hline
\multirow{8}{*}{\begin{tabular}[c]{@{}l@{}}\textbf{Efficient Wait-k}\\\cite{multipath}\end{tabular}} & $k$          & CW    & AP   & AL    & DAL   & BLEU    \\
                                           & 1          & 1.27  & 0.50 & -0.49 & 1.60  & 19.51   \\
                                           & 3          & 1.27  & 0.58 & 1.56  & 3.29  & 24.11   \\
                                           & 5          & 1.39  & 0.66 & 3.71  & 5.18  & 26.85   \\
                                           & 7          & 1.71  & 0.73 & 5.78  & 7.12  & 28.34   \\
                                           & 9          & 2.17  & 0.78 & 7.84  & 8.98  & 29.39   \\
                                           & 11         & 2.78  & 0.82 & 9.73  & 10.79 & 30.02   \\
                                           & 13         & 3.56  & 0.86 & 11.50 & 12.49 & 30.25   \\ \hline
\multirow{15}{*}{\begin{tabular}[c]{@{}l@{}}\textbf{Adaptive Wait-k}\\\cite{zhang-etal-2020-learning-adaptive}\end{tabular}} & ( $\rho_{1}$, $\rho_{13}$ )   & CW    & AP   & AL    & DAL   & BLEU    \\
                                           & (0.02, 0.00)   & 1.54  & 0.54 & 0.83  & 3.27  & 20.29   \\
                                           & (0.04, 0.00)   & 2.07  & 0.56 & 1.40  & 4.59  & 22.34   \\
                                           & (0.05, 0.00)   & 2.28  & 0.58 & 1.90  & 5.25  & 23.56   \\
                                           & (0.06, 0.00)   & 2.58  & 0.60 & 2.43  & 5.99  & 24.59   \\
                                           & (0.07, 0.00)   & 2.79  & 0.62 & 2.94  & 6.57  & 25.96   \\
                                           & (0.09, 0.00)   & 3.25  & 0.66 & 4.10  & 7.78  & 27.44   \\
                                           & (0.10, 0.00)    & 3.45  & 0.68 & 4.66  & 8.31  & 27.88   \\
                                           & (0.10, 0.01) & 3.68  & 0.70 & 5.11  & 8.84  & 28.29   \\
                                           & (0.10, 0.03) & 4.13  & 0.72 & 6.09  & 9.87  & 28.91   \\
                                           & 0.10, 0.05)  & 4.48  & 0.75 & 7.21  & 10.72 & 29.73   \\
                                           & (0.20, 0.00)    & 4.02  & 0.78 & 8.23  & 10.92 & 30.10   \\
                                           & (0.20, 0.05) & 4.75  & 0.82 & 10.12 & 12.35 & 30.76   \\
                                           & (0.20, 0.10)  & 4.68  & 0.85 & 11.55 & 12.98 & 30.78   \\
                                           & (0.30, 0.20)  & 4.16  & 0.86 & 12.18 & 13.09 & 30.74   \\ \hline
\multirow{8}{*}{\begin{tabular}[c]{@{}l@{}}\textbf{MoE Wait-k}\\\cite{zhang-feng-2021-universal}\end{tabular}}       & $k$          & CW    & AP   & AL    & DAL   & BLEU    \\
                                           & 1          & 1.49  & 0.49 & -0.32 & 1.69  & 21.43   \\
                                           & 3          & 1.26  & 0.59 & 1.79  & 3.30  & 25.81   \\
                                           & 5          & 1.37  & 0.66 & 3.88  & 5.18  & 28.34   \\
                                           & 7          & 1.69  & 0.73 & 5.94  & 7.12  & 29.71   \\
                                           & 9          & 2.17  & 0.78 & 7.86  & 8.99  & 30.61   \\
                                           & 11         & 2.78  & 0.82 & 9.73  & 10.78 & 30.89   \\
                                           & 13         & 3.56  & 0.86 & 11.53 & 12.48 & 31.08   \\ \hline
\multirow{6}{*}{\begin{tabular}[c]{@{}l@{}}\textbf{MMA}\\\cite{Ma2019a}\end{tabular}}              & $\lambda$     & CW    & AP   & AL    & DAL   & BLEU    \\
                                           & 0.4        & 2.35  & 0.68 & 4.97  & 7.51  & 28.66   \\
                                           & 0.3        & 2.64  & 0.72 & 6.00  & 9.30  & 29.11   \\
                                           & 0.25       & 3.35  & 0.78 & 8.03  & 12.28 & 28.92   \\
                                           & 0.2        & 4.03  & 0.83 & 9.98  & 14.86 & 28.18   \\
                                           & 0.1        & 14.88 & 0.97 & 13.25 & 19.48 & 27.47   \\ \hline
\multirow{6}{*}{\begin{tabular}[c]{@{}l@{}}\textbf{GMA}\\\cite{gma}\end{tabular}}              & $\delta$     & CW    & AP   & AL    & DAL   & BLEU    \\
                                           & 0.9        & 1.33  & 0.64 & 3.87  & 4.61  & 28.12   \\
                                           & 1.0        & 1.49  & 0.67 & 4.66  & 5.56  & 28.50   \\
                                           & 2.0        & 1.85  & 0.72 & 5.79  & 7.75  & 28.71   \\
                                           & 2.2        & 2.01  & 0.73 & 6.13  & 8.43  & 29.23   \\
                                           & 2.4        & 5.89  & 0.96 & 14.05 & 25.76 & 31.31   \\ \hline
\multirow{5}{*}{\begin{tabular}[c]{@{}l@{}}\textbf{GSiMT}\\\cite{miao-etal-2021-generative}\end{tabular}}            & $\zeta$       & CW    & AP   & AL    & DAL   & BLEU    \\
                                           & 4          & -     & -    & 3.64  & -     & 28.82   \\
                                           & 5          & -     & -    & 4.45  & -     & 29.50   \\
                                           & 6          & -     & -    & 5.13  & -     & 29.78   \\
                                           & 7          & -     & -    & 6.24  & -     & 29.63   \\ \hline
\multirow{16}{*}{\textbf{Wait-info}}       & $\mathcal{K}$          & CW    & AP   & AL    & DAL   & BLEU    \\
                                           & 1          & 1.29  & 0.61 & 3.00  & 3.77  & 27.55   \\
                                           & 2          & 1.36  & 0.64 & 3.78  & 4.56  & 28.89   \\
                                           & 3          & 1.44  & 0.67 & 4.68  & 5.46  & 29.66   \\
                                           & 4          & 1.53  & 0.71 & 5.71  & 6.43  & 30.12   \\
                                           & 5          & 1.68  & 0.74 & 6.66  & 7.37  & 30.59   \\
                                           & 6          & 1.86  & 0.77 & 7.62  & 8.33  & 31.13   \\
                                           & 7          & 2.10  & 0.79 & 8.57  & 9.26  & 31.28   \\
                                           & 8          & 2.38  & 0.81 & 9.48  & 10.18 & 31.39   \\
                                           & 9          & 2.66  & 0.83 & 10.41 & 11.11 & 31.55   \\
                                           & 10         & 3.01  & 0.85 & 11.31 & 11.97 & 31.68   \\
                                           & 11         & 3.38  & 0.87 & 12.16 & 12.82 & 31.66   \\
                                           & 12         & 3.81  & 0.88 & 12.99 & 13.64 & 31.69   \\
                                           & 13         & 4.25  & 0.89 & 13.79 & 14.43 & 31.88   \\
                                           & 14         & 4.73  & 0.90 & 14.56 & 15.19 & 31.94   \\
                                           & 15         & 5.20  & 0.91 & 15.32 & 15.92 & 32.05   \\ \hlinew{1.2pt}
\end{tabular}
\caption{Numerical results on De$\rightarrow$En with Transformer-Base.}
\label{res_deen_base}
\end{table*}

\begin{table*}[t]
\centering
\scriptsize
\begin{tabular}{l|cccccc}  \hlinew{1.2pt}
\multicolumn{7}{c}{\textbf{WMT15 German$\rightarrow$English$\;\;\;\;\;$Transformer-Big}}                                                     \\ \hline
\multirow{2}{*}{\begin{tabular}[c]{@{}l@{}}\textbf{Full-sentence} MT\\\cite{NIPS2017_7181}\end{tabular}} &            & CW    & AP   & AL    & DAL   & BLEU  \\
                                           &            & 27.77 & 1.00 & 27.77 & 27.77 & 32.94 \\ \hline
\multirow{8}{*}{\begin{tabular}[c]{@{}l@{}}\textbf{Wait-k}\\\cite{ma-etal-2019-stacl}\end{tabular}}           & $k$          & CW    & AP   & AL    & DAL   & BLEU  \\
                                           & 1          & 1.16  & 0.52 & 0.25  & 1.82  & 19.13 \\
                                           & 3          & 1.20  & 0.60 & 2.23  & 3.41  & 25.45 \\
                                           & 5          & 1.36  & 0.67 & 4.00  & 5.23  & 28.67 \\
                                           & 7          & 1.70  & 0.73 & 5.97  & 7.17  & 30.12 \\
                                           & 9          & 2.17  & 0.78 & 7.95  & 9.03  & 31.46 \\
                                           & 11         & 2.79  & 0.82 & 9.75  & 10.82 & 31.83 \\
                                           & 13         & 3.56  & 0.86 & 11.59 & 12.51 & 32.08 \\ \hline
\multirow{8}{*}{\begin{tabular}[c]{@{}l@{}}\textbf{Efficient Wait-k}\\\cite{multipath}\end{tabular}} & $k$          & CW    & AP   & AL    & DAL   & BLEU  \\
                                           & 1          & 1.23  & 0.51 & -0.19 & 1.79  & 20.56 \\
                                           & 3          & 1.26  & 0.59 & 1.73  & 3.36  & 25.45 \\
                                           & 5          & 1.39  & 0.66 & 3.82  & 5.24  & 28.58 \\
                                           & 7          & 1.71  & 0.73 & 5.89  & 7.16  & 30.13 \\
                                           & 9          & 2.17  & 0.78 & 7.88  & 9.02  & 31.23 \\
                                           & 11         & 2.78  & 0.82 & 9.77  & 10.81 & 31.52 \\
                                           & 13         & 3.56  & 0.86 & 11.58 & 12.51 & 32.02 \\ \hline
\multirow{14}{*}{\begin{tabular}[c]{@{}l@{}}\textbf{Adaptive Wait-k}\\\cite{zhang-etal-2020-learning-adaptive}\end{tabular}} & ( $\rho_{1}$, $\rho_{13}$ )   & CW    & AP   & AL    & DAL   & BLEU  \\
                                           & (0.02, 0.00)   & 1.42  & 0.54 & 0.99  & 3.00  & 20.50 \\
                                           & (0.04, 0.00)   & 1.86  & 0.56 & 1.37  & 4.22  & 22.62 \\
                                           & (0.05, 0.00)   & 2.10  & 0.57 & 1.69  & 4.81  & 23.77 \\
                                           & (0.06, 0.00)   & 2.36  & 0.59 & 2.23  & 5.54  & 25.43 \\
                                           & (0.07, 0.00)   & 2.58  & 0.61 & 2.70  & 6.14  & 27.06 \\
                                           & (0.08, 0.00)   & 2.84  & 0.63 & 3.17  & 6.75  & 27.96 \\
                                           & (0.09, 0.00)   & 3.08  & 0.65 & 3.72  & 7.33  & 28.92 \\
                                           & (0.10, 0.00)    & 3.28  & 0.67 & 4.28  & 7.88  & 29.90 \\
                                           & (0.10, 0.03) & 3.95  & 0.71 & 5.59  & 9.43  & 30.97 \\
                                           & (0.10, 0.05) & 4.36  & 0.74 & 6.70  & 10.41 & 31.30 \\
                                           & (0.20, 0.00)    & 3.90  & 0.78 & 8.09  & 10.80 & 32.38 \\
                                           & (0.20, 0.05) & 4.78  & 0.82 & 10.00 & 12.35 & 32.46 \\
                                           & (0.30, 0.20)  & 4.16  & 0.86 & 12.19 & 13.11 & 32.24 \\ \hline
\multirow{8}{*}{\begin{tabular}[c]{@{}l@{}}\textbf{MoE Wait-k}\\\cite{zhang-feng-2021-universal}\end{tabular}}       & $k$          & CW    & AP   & AL    & DAL   & BLEU  \\
                                           & 1          & 1.41  & 0.51 & 0.16  & 1.79  & 21.76 \\
                                           & 3          & 1.28  & 0.59 & 2.03  & 3.37  & 26.51 \\
                                           & 5          & 1.37  & 0.67 & 4.03  & 5.22  & 29.33 \\
                                           & 7          & 1.70  & 0.73 & 5.95  & 7.14  & 30.66 \\
                                           & 9          & 2.17  & 0.78 & 7.86  & 8.99  & 30.61 \\
                                           & 11         & 2.78  & 0.82 & 9.73  & 10.78 & 30.89 \\
                                           & 13         & 3.56  & 0.86 & 11.53 & 12.48 & 31.08 \\ \hline
\multirow{9}{*}{\begin{tabular}[c]{@{}l@{}}\textbf{MMA}\\\cite{Ma2019a}\end{tabular}}              & $\lambda$     & CW    & AP   & AL    & DAL   & BLEU  \\
                                           & 1          & 1,69  & 0.56 & 3.00  & 4.03  & 26.10 \\
                                           & 0.75       & 1.66  & 0.58 & 3.40  & 4.46  & 26.50 \\
                                           & 0.5        & 1.69  & 0.59 & 3.69  & 4.83  & 27.70 \\
                                           & 0.4        & 1.70  & 0.59 & 3.75  & 4.90  & 29.20 \\
                                           & 0.3        & 1.82  & 0.60 & 4.18  & 5.35  & 30.30 \\
                                           & 0.27       & 2.37  & 0.71 & 5.91  & 8.27  & 30.88 \\
                                           & 0.25       & 2.62  & 0.75 & 7.02  & 9.88  & 31.04 \\
                                           & 0.2        & 3.21  & 0.79 & 8.75  & 12.60 & 31.08 \\ \hline
\multirow{6}{*}{\begin{tabular}[c]{@{}l@{}}\textbf{GMA}\\\cite{gma}\end{tabular}}              & $\delta$      & CW    & AP   & AL    & DAL   & BLEU  \\
                                           & 1.0        & 1.54  & 0.68 & 4.60  & 5.89  & 30.20 \\
                                           & 2.0        & 1.98  & 0.74 & 6.34  & 8.18  & 30.64 \\
                                           & 2.2        & 2.13  & 0.75 & 6.86  & 8.91  & 31.33 \\
                                           & 2.4        & 2.28  & 0.76 & 7.28  & 9.59  & 31.62 \\
                                           & 2.5        & 3.10  & 0.88 & 12.06 & 20.43 & 31.91 \\ \hline
\multirow{16}{*}{\textbf{Wait-info}}       & $\mathcal{K}$          & CW    & AP   & AL    & DAL   & BLEU  \\
                                           & 1          & 1.30  & 0.62 & 3.41  & 4.17  & 29.19 \\
                                           & 2          & 1.37  & 0.65 & 4.19  & 4.90  & 30.42 \\
                                           & 3          & 1.46  & 0.69 & 5.12  & 5.79  & 31.26 \\
                                           & 4          & 1.56  & 0.72 & 6.05  & 6.74  & 31.68 \\
                                           & 5          & 1.71  & 0.75 & 6.96  & 7.65  & 32.04 \\
                                           & 6          & 1.88  & 0.77 & 7.94  & 8.57  & 32.32 \\
                                           & 7          & 2.14  & 0.80 & 8.83  & 9.49  & 32.56 \\
                                           & 8          & 2.40  & 0.82 & 9.75  & 10.38 & 32.86 \\
                                           & 9          & 2.68  & 0.84 & 10.66 & 11.25 & 32.99 \\
                                           & 10         & 3.00  & 0.85 & 11.53 & 12.13 & 33.10 \\
                                           & 11         & 3.38  & 0.87 & 12.35 & 12.93 & 32.99 \\
                                           & 12         & 3.79  & 0.88 & 13.15 & 13.72 & 33.10 \\
                                           & 13         & 4.21  & 0.89 & 13.94 & 14.48 & 33.23 \\
                                           & 14         & 4.67  & 0.91 & 14.69 & 15.21 & 33.23 \\
                                           & 15         & 5.15  & 0.92 & 15.42 & 15.93 & 33.31 \\ \hlinew{1.2pt}
\end{tabular}
\caption{Numerical results on De$\rightarrow$En with Transformer-Big.}
\label{res_deen_big}
\end{table*}

\end{document}